# Completeness and Performance of the APO Algorithm

**Tal Grinshpoun**  GRINSHPO@CS.BGU.AC.IL
**Amnon Meisels**  AM@CS.BGU.AC.IL
*Ben-Gurion University of the Negev*
*Department of Computer Science*
*Beer-Sheva, Israel*

## Abstract

Asynchronous Partial Overlay (APO) is a search algorithm that uses cooperative mediation to solve Distributed Constraint Satisfaction Problems (DisCSPs). The algorithm partitions the search into different subproblems of the DisCSP. The original proof of completeness of the APO algorithm is based on the growth of the size of the subproblems. The present paper demonstrates that this expected growth of subproblems does not occur in some situations, leading to a termination problem of the algorithm. The problematic parts in the APO algorithm that interfere with its completeness are identified and necessary modifications to the algorithm that fix these problematic parts are given. The resulting version of the algorithm, Complete Asynchronous Partial Overlay (CompAPO), ensures its completeness. Formal proofs for the soundness and completeness of CompAPO are given. A detailed performance evaluation of CompAPO comparing it to other DisCSP algorithms is presented, along with an extensive experimental evaluation of the algorithm's unique behavior. Additionally, an optimization version of the algorithm, CompOptAPO, is presented, discussed, and evaluated.

## 1. Introduction

Algorithms that solve Distributed Constraint Satisfaction Problems (DisCSPs) attempt to achieve concurrency during problem solving in order to utilize the distributed nature of these problems. Distributed backtracking, which forms the majority of DisCSP algorithms, can take many forms. Asynchronous Backtracking (ABT) (Yokoo, Durfee, Ishida, & Kuwabara, 1998; Yokoo & Hirayama, 2000), Asynchronous Forward-Checking (AFC) (Meisels & Zivan, 2007), and Concurrent Dynamic Backtracking (ConcDB) (Zivan & Meisels, 2006a) are representative examples of the family of distributed backtracking algorithms. All of these algorithms maintain one or more partial solutions of the DisCSP and attempt to extend the partial solution into a complete one. The ABT algorithm attempts to achieve concurrency by asynchronously assigning values to the variables. The AFC algorithm performs value assignments synchronously, but achieves its concurrency by performing asynchronous computation in the form of forward checking. The ConcDB algorithm concurrently attempts to extend multiple partial solutions, scanning different parts of the search space.

A completely different approach to achieve concurrency can be by the merging of partial solutions into a complete one. The inherent concurrency of merging partial solutions makes it a fascinating paradigm for solving DisCSPs. However, such an approach is prone to many errors – deadlocks could prevent termination, and failures could occur in the attempt to merge all of the partial solutions. Consequently, it is hard to develop such an algorithm that is both correct and well performing. A recently published algorithm, Asynchronous Partial Overlay (APO) (Mailler, 2004; Mailler & Lesser, 2006), attempts to solve DisCSPs by merging partial solutions. It uses the concept of mediation to centralize the search procedure in different parts of the DisCSP. Due to its unique approach, several researchers have already proposed changes and modifications to the APO algorithm (Benisch & Sadeh, 2006; Semnani & Zamanifar, 2007). Unfortunately, none of these studies has examined the completeness of APO. Additionally, the distinctive behavior of the APO algorithm calls for a thorough experimental evaluation. The present paper presents an in-depth investigation of the completeness and termination of the APO algorithm, constructs a correct version of the algorithm – CompAPO – and goes on to present extensive experimental evaluation of the complete APO algorithm.

The APO algorithm partitions the agents into groups that attempt to find consistent partial solutions. The partition mechanism is dynamic during search and enables a dynamic change of groups. The key factor in the termination (and consequently the completeness) of the APO algorithm as presented in the original correctness proof (Mailler & Lesser, 2006) is the monotonic growth of initially partitioned groups during search. This growth arises because the subproblems overlap, allowing agents to increase the size of the subproblems they solve. We have discovered that this expected growth of groups does not occur in some situations, leading to a termination problem of the APO algorithm. Nevertheless, the unique way in which APO attempts to solve DisCSPs has encouraged us to try and fix it.

The termination problem of the APO algorithm is shown in section 4 by constructing a scenario that leads to an infinite loop of the algorithm's run (Grinshpoun, Zazon, Binshtok, & Meisels, 2007). Such a running example is essential to the understanding of APO's completeness problem, since the algorithm is very complex. To help understand the problem, a full pseudo-code of APO that follows closely the original presentation of the algorithm (Mailler & Lesser, 2006) is given. The erroneous part in the proof of APO's completeness as presented by Mailler and Lesser (2006) is shown and the problematic parts in the algorithm that interfere with its completeness are identified. Necessary modifications to the algorithm are proposed, in order to fix these problematic parts. The resulting version of the algorithm ensures its completeness, and is termed Complete Asynchronous Partial Overlay (CompAPO) (Grinshpoun & Meisels, 2007). Formal proofs for the soundness and completeness of CompAPO are presented.

The modifications of CompAPO may potentially affect the performance of the algorithm. Also, in the evaluation of the original APO algorithm (Mailler & Lesser, 2006), it was compared to the AWC algorithm (Yokoo, 1995), which is not an efficient DisCSP solver (Zivan, Zazone, & Meisels, 2007). Moreover, the tests in the work of Mailler and Lesser (2006) were only performed on relatively sparse problems, and the comparison with AWC was made by the use of some problematic measures. An extensive experimental evaluation of CompAPO compares its performance with other DisCSP search algorithms on randomly generated DisCSPs. Our experiments show that CompAPO performs significantly different than other DisCSP algorithms, which is not surprising considering its singular way of problem solving.

Asynchronous Partial Overlay is actually a family of algorithms. The completeness and termination problems that are presented and corrected in the present study apply to all the members of the family. The OptAPO algorithm (Mailler & Lesser, 2004; Mailler, 2004) is an optimization version of APO that solves Distributed Constraint Optimization Problems (DisCOPs). The present paper proposes similar modifications to those of APO in order to achieve completeness for OptAPO. The resulting CompOptAPO algorithm is evaluated extensively on randomly generated DisCOPs.

The plan of the paper is as follows. DisCSPs are presented briefly in section 2. Section 3, gives a short description of the APO algorithm along with its pseudo-code. An infinite loop scenario for APO is described in detail in section 4 and the problems that lead to the infinite looping are analyzed in section 5. Section 6 presents a detailed solution to the problem that forms the CompAPO version of the algorithm, followed by proofs for the soundness and completeness of CompAPO (section 7). An optimization version of the algorithm, CompOptAPO, is presented and discussed in section 8. An extensive performance evaluation of CompAPO and CompOptAPO is in section 9. Our conclusions are summarized in section 10.

## 2. Distributed Constraint Satisfaction

A distributed constraints satisfaction problem – DisCSP, is composed of a set of $k$ agents $A_1, A_2, ..., A_k$. Each agent $A_i$ contains a set of constrained variables $x_{i_1}, x_{i_2}, ..., x_{i_{n_i}}$. Constraints or **relations** $R$ are subsets of the Cartesian product of the domains of the constrained variables. For a set of constrained variables $x_{i_k}, x_{j_l}, ..., x_{m_n}$, with domains of values for each variable $D_{i_k}, D_{j_l}, ..., D_{m_n}$, the constraint is defined as $R \subseteq D_{i_k} \times D_{j_l} \times ... \times D_{m_n}$. A **binary constraint** $R_{ij}$ between any two variables $x_j$ and $x_i$ is a subset of the Cartesian product of their domains – $R_{ij} \subseteq D_j \times D_i$. In a distributed constraint satisfaction problem (DisCSP), the agents are connected by constraints between variables that belong to different agents (Yokoo et al., 1998). In addition, each agent has a set of constrained variables, i.e. a *local constraint network*.

An assignment (or a label) is a pair $< var, val >$, where *var* is a variable of some agent and *val* is a value from *var*'s domain that is assigned to it. A *compound label* is a set of assignments of values to a set of variables. A **solution** $s$ to a DisCSP is a compound label that includes all variables of all agents, which satisfies all the constraints. Agents check assignments of values against non-local constraints by communicating with other agents through sending and receiving messages.

Current studies of DisCSPs follow the assumption that all agents hold exactly one variable (Yokoo & Hirayama, 2000; Bessiere, Maestre, Brito, & Meseguer, 2005). Accordingly, the present study often uses the variable's name $x_i$ to represent the agent it belongs to ($A_i$). In addition, the following common assumptions are used in the present study:

- The amount of time that passes between the sending and the receiving of a message is finite.
- Messages sent by agent $A_i$ to agent $A_j$ are received by $A_j$ in the order they were sent.

## 3. Asynchronous Partial Overlay

Asynchronous Partial Overlay (APO) is an algorithm for solving DisCSPs that applies cooperative mediation. The pseudo-code in Algorithms 1, 2, and 3 follows closely the presentation of APO in the work of Mailler and Lesser (2006).

**Algorithm 1** APO procedures for initialization and local resolution.

**procedure initialize**
1: $d_i \leftarrow random\ d \in D_i$;
2: $p_i \leftarrow sizeof(neighbors) + 1$;
3: $m_i \leftarrow$ **true**;
4: $mediate \leftarrow$ **false**;
5: add $x_i$ to the *good_list*;
6: send (**init**, $(x_i, p_i, d_i, m_i, D_i, C_i)$) to neighbors;
7: $init\_list \leftarrow$ neighbors;

**when received** (**init**, $(x_j, p_j, d_j, m_j, D_j, C_j)$) **do**
1: add $(x_j, p_j, d_j, m_j, D_j, C_j)$ to *agent_view*;
2: **if** $x_j$ is a neighbor of some $x_k \in good\_list$ **do**
3:     add $x_j$ to the *good_list*;
4:     add all $x_l \in agent\_view \land x_l \notin good\_list$ that can now be connected to the *good_list*;
5:     $p_i \leftarrow sizeof(good\_list)$;
6: **if** $x_j \notin init\_list$ **do**
7:     send (**init**, $(x_i, p_i, d_i, m_i, D_i, C_i)$) to $x_j$;
8: **else**
9:     remove $x_j$ from *init_list*;
10: **check_agent_view**;

**when received** (**ok?**, $(x_j, p_j, d_j, m_j)$) **do**
1: update *agent_view* with $(x_j, p_j, d_j, m_j)$;
2: **check_agent_view**;

**procedure check_agent_view**
1: **if** $init\_list \neq \emptyset$ or $mediate \neq$ **false do**
2:     **return**;
3: $m'_i \leftarrow hasConflict(x_i)$;
4: **if** $m'_i$ and $\neg \exists j(p_j > p_i \land m_j ==$ **true**) **do**
5:     **if** $\exists(d'_i \in D_i)(d'_i \cup agent\_view$ does not conflict) **and** $d_i$ conflicts exclusively with lower priority neighbors **do**
6:        $d_i \leftarrow d'_i$;
7:        send (**ok?**, $(x_i, p_i, d_i, m_i)$) to all $x_j \in agent\_view$;
8:     **else**
9:        **mediate**;
10: **else if** $m_i \neq m'_i$ **do**
11:     $m_i \leftarrow m'_i$;
12:     send (**ok?**, $(x_i, p_i, d_i, m_i)$) to all $x_j \in agent\_view$;

**Algorithm 2** Procedures for mediating an APO session and for choosing a solution during an APO mediation.

**procedure mediate**
1: $preferences \leftarrow \emptyset$;
2: $counter \leftarrow 0$;
3: **for each** $x_j \in good\_list$ **do**
4:    send (**evaluate?**, $(x_i, p_i)$) to $x_j$;
5:    $counter \leftarrow counter + 1$;
6: $mediate \leftarrow$ **true**;

**when received** (**wait!**, $(x_j, p_j)$) **do**
1: update $agent\_view$ with $(x_j, p_j)$;
2: $counter \leftarrow counter - 1$;
3: **if** $counter == 0$ **do choose_solution**;

**when received** (**evaluate!**, $(x_j, p_j, labeled\ D_j)$) **do**
1: record $(x_j, labeled\ D_j)$ in $preferences$;
2: update $agent\_view$ with $(x_j, p_j)$;
3: $counter \leftarrow counter - 1$;
4: **if** $counter == 0$ **do choose_solution**;

**procedure choose_solution**
1: select a solution $s$ using a Branch and Bound search that:
2:    1. satisfies the constraints between agents in the $good\_list$
3:    2. minimizes the violations for agents outside of the session
4: **if** $\neg \exists s$ that satisfies the constraints **do**
5:    **broadcast no solution**;
6: **for each** $x_j \in agent\_view$ **do**
7:    **if** $x_j \in preferences$ **do**
8:      **if** $d'_j \in s$ violates an $x_k$ **and** $x_k \notin agent\_view$ **do**
9:        send (**init**, $(x_i, p_i, d_i, m_i, D_i, C_i)$) to $x_k$;
10:       add $x_k$ to $init\_list$;
11:      send (**accept!**, $(d'_j, x_i, p_i, d_i, m_i)$) to $x_j$;
12:      update $agent\_view$ for $x_j$;
13:    **else**
14:      send (**ok?**, $(x_i, p_i, d_i, m_i)$) to $x_j$;
15: $mediate \leftarrow$ **false**;
16: **check_agent_view**;

**Algorithm 3** Procedures for receiving an APO session.

**when received (evaluate?, $(x_j, p_j)$) do**
1: $m_j \leftarrow$ **true**;
2: **if** $mediate ==$ **true** or $\exists k(p_k > p_j \wedge m_k ==$ **true**) **do**
3:    send (**wait!**, $(x_i, p_i)$) to $x_j$;
4: **else**
5:    $mediate \leftarrow$ **true**;
6:    label each $d \in D_i$ with the names of the agents that would be violated by setting $d_i \leftarrow d$;
7:    send (**evaluate!**, $(x_i, p_i, labeled\ D_i)$) to $x_j$;

**when received (accept!, $(d, x_j, p_j, d_j, m_j)$) do**
1: $d_i \leftarrow d$;
2: $mediate \leftarrow$ **false**;
3: send (**ok?**, $(x_i, p_i, d_i, m_i)$) to all $x_j \in agent\_view$;
4: update $agent\_view$ with $(x_j, p_j, d_j, m_j)$;
5: **check_agent_view**;

At the beginning of its problem solving, the APO algorithm performs an initialization phase, in which neighboring agents exchange data through **init** messages (procedure **initialize** in Algorithm 1). Following that, agents check their *agent_view* to identify conflicts between themselves and their neighbors (procedure **check_agent_view** in Algorithm 1). If during this check, an agent finds a conflict with one of its neighbors, it expresses desire to act as a mediator. In case the agent does not have any neighbors that wish to mediate and have a wider view of the constraint graph than itself, the agent successfully assumes the role of mediator.

Using mediation (Algorithms 2, and 3), agents can solve subproblems of the DisCSP by conducting an internal Branch and Bound search (procedure **choose_solution** in Algorithm 2). For a complete solution of the DisCSP, the solutions of the subproblems must be compatible. When solutions of overlapping subproblems have conflicts, the solving agents increase the size of the subproblems that they work on. The original paper (Mailler & Lesser, 2006) uses the term *preferences* to describe potential conflicts between solutions of overlapping subproblems. In the present paper we use the term *external constraints* to describe such conflicts. A detailed description of the APO algorithm can be found in the work of Mailler and Lesser (2006).

## 4. An Infinite Loop Scenario

Consider the 3-coloring problem presented in Figure 1 by the solid lines. Each agent can assign one of the three available colors Red, Green, or Blue. To the standard inequality constraints that the solid lines represent, four weaker constraints (diagonal dashed lines) are added. The dashed lines represent constraints that do not allow only the combinations (Green,Green) and (Blue,Blue) to be assigned by the agents. Ties in the priorities of agents are broken using an anti-lexicographic ordering of their names.

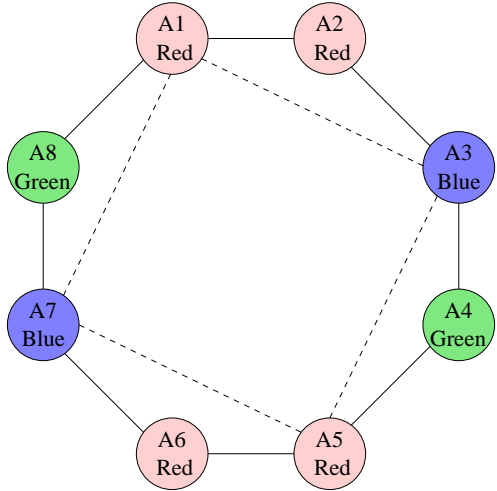

Figure 1: The constraints graph with the initial assignments.

| Agent | Color | $m_i$ | $d_j$ values | $m_j$ values |
|---|---|---|---|---|
| $A_1$ | R | $m_1 = t$ | $d_2$ =R, $d_3$ =B, $d_7$ =B, $d_8$ =G | $m_2 = t$, $m_3 = f$, $m_7 = f$, $m_8 = f$ |
| $A_2$ | R | $m_2 = t$ | $d_1$ =R, $d_3$ =B | $m_1 = t$, $m_3 = f$ |
| $A_3$ | B | $m_3 = f$ | $d_1$ =R, $d_2$ =R, $d_4$ =G, $d_5$ =R | $m_1 = t$, $m_2 = t$, $m_4 = f$, $m_5 = t$ |
| $A_4$ | G | $m_4 = f$ | $d_3$ =B, $d_5$ =R | $m_3 = f$, $m_5 = t$ |
| $A_5$ | R | $m_5 = t$ | $d_3$ =B, $d_4$ =G, $d_6$ =R, $d_7$ =B | $m_3 = f$, $m_4 = f$, $m_6 = t$, $m_7 = f$ |
| $A_6$ | R | $m_6 = t$ | $d_5$ =R, $d_7$ =B | $m_5 = t$, $m_7 = f$ |
| $A_7$ | B | $m_7 = f$ | $d_1$ =R, $d_5$ =R, $d_6$ =R, $d_8$ =G | $m_1 = t$, $m_5 = t$, $m_6 = t$, $m_8 = f$ |
| $A_8$ | G | $m_8 = f$ | $d_1$ =R, $d_7$ =B | $m_1 = t$, $m_7 = f$ |

Table 1: Configuration 1.

The initial selection of values by all agents is depicted in Figure 1. In the initial state, two constraints are violated – $(A_1, A_2)$ and $(A_5, A_6)$. Assume that agents $A_3$, $A_4$, $A_7$, and $A_8$ are the first to complete their initialization phase by exchanging **init** messages with all their neighbors (procedure **initialize** in Algorithm 1). These agents do not have conflicts, therefore they set $m_i \leftarrow$ ***false*** and send **ok?** messages to their neighbors when each of them runs the **check_agent_view** procedure (Algorithm 1). Only after the arrival of the **ok?** messages from agents $A_3$, $A_4$, $A_7$, and $A_8$, do agents $A_1$, $A_2$, $A_5$, and $A_6$ accept the last **init** messages from their other neighbors and complete the initialization phase. Agents $A_2$ and $A_6$ have conflicts, but they complete the **check_agent_view** procedure without mediating or changing their state. This is true, because in the agent views of $A_2$ and $A_6$, $m_1 =$ **true** and $m_5 =$ **true**, respectively. These neighbors have higher priority than agents $A_2$ and $A_6$ respectively. We denote by *configuration 1* the states of all the agents at this point of the processing and present the configuration in Table 1.

After all agents complete their initializations, agents $A_1$ and $A_5$ detect that they have conflicts, and that they have no neighbor with a higher priority that wants to mediate. Consequently, agents $A_1$ and $A_5$ start mediation sessions, since they cannot change their own color to a consistent state with their neighbors.

| Agent | Color | $m_i$ | $d_j$ values | $m_j$ values |
|---|---|---|---|---|
| $A_1$ | G | $m_1 = f$ | $d_2 =$B, $d_3 =$R, $d_7 =$B, $d_8 =$R | $m_2 = f, m_3 = f, m_7 = f, m_8 = f$ |
| $A_2$ | B | $m_2 = f$ | $d_1 =$G, $d_3 =$R | $m_1 = f, m_3 = f$ |
| $A_3$ | R | $m_3 = f$ | $d_1 =$G, $d_2 =$B, $d_4 =$G, $d_5 =$R | $m_1 = f, m_2 = f, m_4 = f, m_5 = t$ |
| $A_4$ | G | $m_4 = f$ | $\boldsymbol{d_3 =}$**B**, $d_5 =$R | $m_3 = f, m_5 = t$ |
| $A_5$ | R | $m_5 = t$ | $\boldsymbol{d_3 =}$**B**, $d_4 =$G, $d_6 =$R, $d_7 =$B | $m_3 = f, m_4 = f, m_6 = t, m_7 = f$ |
| $A_6$ | R | $m_6 = t$ | $d_5 =$R, $d_7 =$B | $m_5 = t, m_7 = f$ |
| $A_7$ | B | $m_7 = f$ | $d_1 =$G, $d_5 =$R, $d_6 =$R, $\boldsymbol{d_8 =}$**G** | $m_1 = f, m_5 = t, m_6 = t, m_8 = f$ |
| $A_8$ | R | $m_8 = f$ | $d_1 =$G, $d_7 =$B | $m_1 = f, m_7 = f$ |

Table 2: Configuration 2 – obsolete data in *agent_views* is in bold face.

Let us first observe $A_1$'s mediation session. $A_1$ sends **evaluate?** messages to its neighbors $A_2$, $A_3$, $A_7$, and $A_8$ (procedure **mediate** in Algorithm 2). All these agents reply with **evaluate!** messages (Algorithm 3). $A_1$ conducts a Branch and Bound search to find a solution that satisfies all the constraints between $A_1$, $A_2$, $A_3$, $A_7$, and $A_8$, and also minimizes external constraints (procedure **choose_solution** in Algorithm 2). In our example, $A_1$ finds the solution ($A_1 \leftarrow$Green, $A_2 \leftarrow$Blue, $A_3 \leftarrow$Red, $A_7 \leftarrow$Blue, $A_8 \leftarrow$Red), which satisfies the internal constraints, and minimizes to zero the external constraints. $A_1$ sends **accept!** messages to its neighbors, informing them of its solution. $A_2$, $A_3$, $A_7$, and $A_8$ receive the **accept!** messages and send **ok?** messages with their new states to their neighbors (Algorithm 3). However, the **ok?** messages from $A_8$ to $A_7$ and from $A_3$ to $A_4$ and to $A_5$ are delayed. Observe that the algorithm is asynchronous and naturally deals with such scenarios.

Concurrently with the above mediation session of $A_1$, agent $A_5$ starts its own mediation session. $A_5$ sends **evaluate?** messages to its neighbors $A_3$, $A_4$, $A_6$, and $A_7$. Let us assume that the message to $A_7$ is delayed. $A_4$ and $A_6$ receive the **evaluate?** messages and reply with **evaluate!**, since they do not know any agents of higher priority than $A_5$ that want to mediate. $A_3$, is in $A_1$'s mediation session, so it replies with **wait!**. We denote by *configuration 2* the states of all the agents at this point of the processing (see Table 2).

Only after $A_1$'s mediation session is over, $A_7$ receives the delayed **evaluate?** message from $A_5$. Since $A_7$ is no longer in a mediation session, nor does it expect a mediation session from a node of higher priority than $A_5$ (see $A_7$'s view in Table 2), agent $A_7$ replies with **evaluate!**. Notice that $A_7$'s view of $d_8$ is obsolete (the **ok?** message from $A_8$ to $A_7$ is still delayed). When agent $A_5$ receives the **evaluate!** message from $A_7$, it can continue the mediation session involving agents $A_4$, $A_5$, $A_6$, and $A_7$. Since the **ok?** messages from $A_3$ to $A_4$ and $A_5$ are also delayed, agent $A_5$ starts its mediation session with knowledge about agents $A_3$ and $A_8$ that is not updated (see bold-faced data in Table 2).

Agent $A_5$ conducts a Branch and Bound search to find a solution that satisfies all the constraints between $A_4$, $A_5$, $A_6$, and $A_7$, that also minimizes external constraints. In our example, $A_5$ finds the solution ($A_4 \leftarrow$Red, $A_5 \leftarrow$Green, $A_6 \leftarrow$Blue, $A_7 \leftarrow$Red), which satisfies the internal constraints, and minimizes to zero the external constraints (remember that $A_5$ has wrong data about the assignments of $A_3$ and $A_8$). $A_5$ sends **accept!** messages to $A_4$, $A_6$, and $A_7$, informing them of its solution. The agents receive these messages and send **ok?** messages with their new states to their neighbors. By now, all the delayed messages get to their destinations, and two constraints are violated – ($A_3$,$A_4$) and ($A_7$,$A_8$). Consequently, agents $A_3$, $A_4$, $A_7$, and $A_8$ want to mediate, whereas agents $A_1$, $A_2$, $A_5$, and $A_6$ do not wish to mediate, since they do not have any conflicts. We denote by *configuration 3* the states of all the agents after $A_5$'s solution has been assigned and all delayed messages arrived at their destinations (see Figure 2 and Table 3).

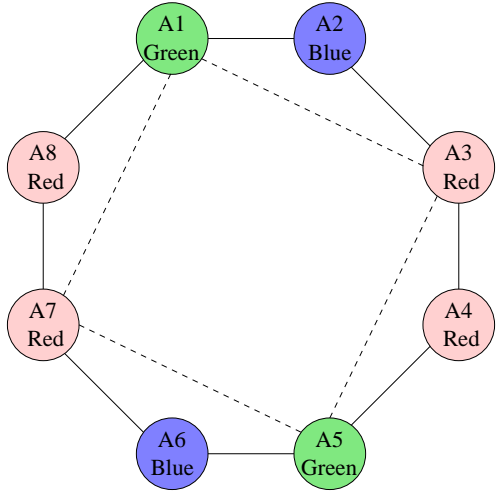

Figure 2: The graph in configuration 3.

| Agent | Color | $m_i$ | $d_j$ values | $m_j$ values |
|---|---|---|---|---|
| $A_1$ | G | $m_1 = f$ | $d_2$ =B, $d_3$ =R, $d_7$ =R, $d_8$ =R | $m_2 = f$, $m_3 = t$, $m_7 = t$, $m_8 = t$ |
| $A_2$ | B | $m_2 = f$ | $d_1$ =G, $d_3$ =R | $m_1 = f$, $m_3 = t$ |
| $A_3$ | R | $m_3 = t$ | $d_1$ =G, $d_2$ =B, $d_4$ =R, $d_5$ =G | $m_1 = f$, $m_2 = f$, $m_4 = t$, $m_5 = f$ |
| $A_4$ | R | $m_4 = t$ | $d_3$ =R, $d_5$ =G | $m_3 = t$, $m_5 = f$ |
| $A_5$ | G | $m_5 = f$ | $d_3$ =R, $d_4$ =R, $d_6$ =B, $d_7$ =R | $m_3 = t$, $m_4 = t$, $m_6 = f$, $m_7 = t$ |
| $A_6$ | B | $m_6 = f$ | $d_5$ =G, $d_7$ =R | $m_5 = f$, $m_7 = t$ |
| $A_7$ | R | $m_7 = t$ | $d_1$ =G, $d_5$ =G, $d_6$ =B, $d_8$ =R | $m_1 = f$, $m_5 = f$, $m_6 = f$, $m_8 = t$ |
| $A_8$ | R | $m_8 = t$ | $d_1$ =G, $d_7$ =R | $m_1 = f$, $m_7 = t$ |

Table 3: Configuration 3.

Until now, we have shown a series of steps that led from *configuration 1* to *configuration 3*. A careful look at Figures 1 and 2 reveals that these configurations are actually isomorphic. Consequently, we will next show a very similar series of steps that will lead us right back to *configuration 1*.

Agents $A_3$ and $A_7$ detect that they have conflicts and that they have no neighbor with a higher priority that wants to mediate. Consequently, agents $A_3$ and $A_7$ start mediation sessions, since they cannot change their own color to a consistent state with their neighbors.

We will first observe $A_3$'s mediation session. $A_3$ sends **evaluate?** messages to its neighbors $A_1$, $A_2$, $A_4$, and $A_5$. All these agents reply with **evaluate!** messages. $A_3$ conducts a Branch and Bound search to find a solution that satisfies all the constraints between $A_1$, $A_2$, $A_3$, $A_4$, and $A_5$, and also minimizes external constraints. Agent $A_3$ finds the solution ($A_1$ ←Green, $A_2$ ←Red, $A_3$ ←Blue, $A_4$ ←Green, $A_5$ ←Red), which satisfies the internal constraints, and minimizes to zero the external constraints. $A_3$ sends **accept!** messages to its neighbors, informing them of its solution. $A_1$, $A_2$, $A_4$, and $A_5$ receive the **accept!** messages and send **ok?** messages with their new states to their neighbors. However, the **ok?** messages from $A_2$ to $A_1$ and from $A_5$ to $A_6$ and to $A_7$ are delayed.

| Agent | Color | $m_i$ | $d_j$ values | $m_j$ values |
|-------|-------|-------|--------------|--------------|
| $A_1$ | G | $m_1 = f$ | $\boldsymbol{d_2} =$**B**, $d_3$ =B, $d_7$ =R, $d_8$ =R | $m_2 = f$, $m_3 = f$, $m_7 = t$, $m_8 = t$ |
| $A_2$ | R | $m_2 = f$ | $d_1$ =G, $d_3$ =B | $m_1 = f$, $m_3 = f$ |
| $A_3$ | B | $m_3 = f$ | $d_1$ =G, $d_2$ =R, $d_4$ =G, $d_5$ =R | $m_1 = f$, $m_2 = f$, $m_4 = f$, $m_5 = f$ |
| $A_4$ | G | $m_4 = f$ | $d_3$ =B, $d_5$ =R | $m_3 = f$, $m_5 = f$ |
| $A_5$ | R | $m_5 = f$ | $d_3$ =B, $d_4$ =G, $d_6$ =B, $d_7$ =R | $m_3 = f$, $m_4 = f$, $m_6 = f$, $m_7 = t$ |
| $A_6$ | B | $m_6 = f$ | $\boldsymbol{d_5} =$**G**, $d_7$ =R | $m_5 = f$, $m_7 = t$ |
| $A_7$ | B | $m_7 = t$ | $d_1$ =G, $\boldsymbol{d_5} =$**G**, $d_6$ =B, $d_8$ =R | $m_1 = f$, $m_5 = f$, $m_6 = f$, $m_8 = t$ |
| $A_8$ | R | $m_8 = t$ | $d_1$ =G, $d_7$ =R | $m_1 = f$, $m_7 = t$ |

Table 4: Configuration 4 – obsolete data in *agent_views* is in bold face.

Concurrently with the above mediation session of $A_3$, agent $A_7$ starts its own mediation session. $A_7$ sends **evaluate?** messages to its neighbors $A_1$, $A_5$, $A_6$, and $A_8$. Let us assume that the message to $A_1$ is delayed. $A_6$ and $A_8$ receive the **evaluate?** messages and reply with **evaluate!**, since they do not know any agents of higher priority than $A_7$ that want to mediate. $A_5$, is in $A_3$'s mediation session, so it replies with **wait!**. We denote by *configuration 4* the states of all the agents at this point of the processing (see Table 4).

Only after $A_3$'s mediation session is over, $A_1$ receives the delayed **evaluate?** message from $A_7$. Since $A_1$ is no longer in a mediation session, nor does it expect a mediation session from a node of higher priority than $A_7$ (see $A_1$'s view in Table 4), agent $A_1$ replies with **evaluate!**. Notice that $A_1$'s view of $d_2$ is obsolete (the **ok?** message from $A_2$ to $A_1$ is still delayed). When agent $A_7$ receives the **evaluate!** message from $A_1$, it can continue the mediation session involving agents $A_1$, $A_6$, $A_7$, and $A_8$. Since the **ok?** messages from $A_5$ to $A_6$ and $A_7$ are also delayed, agent $A_7$ starts its mediation session with knowledge about agents $A_2$ and $A_5$ that is not updated (see bold-faced data in Table 4).

Agent $A_7$ conducts a Branch and Bound search to find a solution that satisfies all the constraints between $A_1$, $A_6$, $A_7$, and $A_8$, that also minimizes external constraints. In our example, $A_7$ finds the solution ($A_1$ ←Red, $A_6$ ←Red, $A_7$ ←Blue, $A_8$ ←Green), which satisfies the internal constraints, and minimizes to zero the external constraints (remember that $A_7$ has wrong data about $A_2$ and $A_5$). $A_7$ sends **accept!** messages to $A_1$, $A_6$, and $A_8$, informing them of its solution. The agents receive these messages and send **ok?** messages with their new states to their neighbors. By now, all the delayed messages get to their destination, and two constraints are violated – ($A_1$,$A_2$) and ($A_5$,$A_6$). Consequently, agents $A_1$, $A_2$, $A_5$, and $A_6$ want to mediate, whereas agents $A_3$, $A_4$, $A_7$, and $A_8$ do not wish to mediate, since they do not have any conflicts. Notice that all the agents have returned to the exact states they were in *configuration 1* (see Figure 1 and Table 1).

The cycle that we have just shown between *configuration 1* and *configuration 3* can continue indefinitely. This example contradicts the termination and completeness of the APO algorithm.

It should be noted that we did not mention all the messages passed in the running of our example. We mentioned only those messages that are important for the understanding of the example, since the example is complicated enough. For instance, after agent $A_1$ completes its mediation session (before *configuration 2*), there is some straightforward exchange of messages between agents, before the $m_j$ values of all the agents become correct (as presented in Table 2).

## 5. Analyzing the Problems

In the previous section a termination problem of the APO algorithm was described by constructing a scenario that leads to an infinite loop of the algorithm's run. To better understand the completeness problem of APO, one must refer to the completeness proof of the APO algorithm as given by Mailler and Lesser (2006). The proof is based on the incorrect assertion that when a mediation session terminates it has three possible outcomes:

1. A solution with no external conflicts.
2. No solution exists.
3. A solution with at least one external violated constraint.

In the first case, the mediator presumably finds a solution to the subproblem. In the second case, the mediator discovers that the overall problem is unsolvable. In the third case, the mediator adds the agent (or agents) with whom the external conflicts were found to its *good_list*, which is used to define future mediations. In this way, either a solution or no solution is found (first two cases), or the *good_list* grows, consequently bringing the problem solving closer to a centralized solution (third case).

However, the infinite loop scenario in section 4 shows that the assertion claiming that these three cases cover all the possible outcomes of a mediation session is incorrect. There are two possible reasons for this incorrectness. The first reason is the possibility that a mediator initiates a partial mediation session without obtaining a lock on all the agents in its *good_list*. The second reason is incorrect information about external constraints when neighboring mediation sessions are performed concurrently. Both reasons relate to the concurrency of mediation sessions.

### 5.1 Partial Mediation Sessions

The first reason for the incorrectness of the "always growth" assertion is the possibility that a mediator initiates a partial mediation session without obtaining a lock on all the agents in its *good_list*. This possibility can occur because of earlier engagements of some of its *good_list*'s members with other mediation sessions. In APO's code, these agents send a **wait!** message.

Let us consider some partial mediation session. Assume that the mediator finds a solution to the subproblem, but such that has external conflicts with agents outside the mediation session. Assume also that all these conflicts are with agents that are already in the mediator's *good_list*. Notice that this is possible, since these agents can be engaged in other mediation sessions and have earlier sent **wait!** messages to the mediator. The present mediation session falls into case 3 of the original proof. However, it is apparent that no new agents will be added to the *good_list* – contradicting the assertion.

Another possible outcome of partial mediation sessions is a situation in which an agent or several agents that have the entire graph in their *good_list* try to mediate, but fail to get a lock on all the agents in their *good_list*. Consequently, the situation in which a single agent holds the entire constraint graph, does not necessarily lead to a solution, due to an oscillation.

### 5.2 Neighboring Mediation Sessions

The second reason for the incorrectness of the assertion in the original proof (Mailler & Lesser, 2006) is the potential existence of obsolete information of external constraints. This reason involves a scenario in which two neighboring mediation sessions are performed concurrently. Both the mediation sessions in the scenario end with finding a solution that presumably has no external conflicts, but the combination of both solutions causes new conflicts. This was the case with the mediation sessions of agents $A_1$ and $A_5$ in the example of section 4. Such a scenario seemingly fits the first case in the assertion, in which no external conflicts are found by each of the mediation sessions. Consequently, no external-conflict-free partial solution is found – contradicting the assertion. Furthermore, none of the mediators increase their *good_list*. This enables the occurrence of an infinite loop, as displayed in section 4.

## 6. Complete Asynchronous Partial Overlay

A two-part solution that solves the completeness problem of APO is presented. The first part of the solution insures that the first reason for the incorrectness of the assertion (see section 5.1) could not occur. This is achieved by preventing partial mediation sessions that go on without the participation of the entire mediator's *good_list*. The second part of the solution addresses the scenario in which two neighboring mediation sessions are performed concurrently. In such a scenario, the results of the mediation sessions can create new conflicts. In order to ensure that *good_lists* grow and rule out an infinite loop, the second part of the solution makes sure that at least one of the *good_lists* grows. Combined with the first part that insures that mediation sessions will involve the entire *good_lists* of the mediators, the completeness of APO is secured.

### 6.1 Preventing Partial Mediation Sessions

Our proposed algorithm disables the initiation of partial mediation sessions by making the mediator wait until it obtains a lock on all the agents in its *good_list*. Algorithm 4 presents the changes and additions to APO that are needed for preventing partial mediation sessions.

When the mediator receives a **wait!** message from at least one of the agents in its *good_list*, it simply cancels the mediation session (**wait!**, line 2) and sets the *counter* to a special value of -1 (**wait!**, line 3). To notify the other participants of the canceled mediation session, the mediator sends a **cancel!** message to each of the participants (**wait!**, line 4). Upon receiving a **cancel!** message, the receiving agent updates its *agent_view* (**cancel!**, line 1) and frees itself from the mediator's lock (**cancel!**, line 2). However, the agent is still

**Algorithm 4** Preventing partial mediation sessions.
**when received** (**wait!**, $(x_j, p_j)$) **do**
1: update *agent_view* with $(x_j, p_j)$;
2: *mediate* ← **false**;
3: *counter* ← −1;
4: send (**cancel!**, $(x_i, p_i)$) to all $x_j \in good\_list$;
5: **check_agent_view**;

**when received** (**evaluate!**, $(x_j, p_j, labeledD_j)$) **do**
1: update *agent_view* with $(x_j, p_j)$;
2: **if** *counter* ≠ −1 **do**
3:    record $(x_j, labeledD_j)$ in *preferences*;
4:    *counter* ← *counter* − 1;
5:    **if** *counter* = 0 **do choose_solution**;

**when received** (**cancel!**, $(x_j, p_j)$) **do**
1: update *agent_view* with $(x_j, p_j)$;
2: *mediate* ← **false**;
3: **check_agent_view**;

aware of the mediator's willingness to mediate. Consequently, it will not join a mediation session of a lower priority agent. The special value of *counter* is used by the mediator to disregard **evaluate!** messages that arrive *after* a **wait!** message (that causes a cancellation) due to asynchronous message passing (**evaluate!**, line 2).

The cancellation of a mediation session upon receiving a single **wait!** message introduces a need for a unique identification for mediation sessions. Consider a **wait!** message that a mediator receives. Upon receiving the message, the mediator cancels the mediation session and calls **check_agent_view**. It may decide to initiate a new mediation session. However, it might receive a **wait!** message from another agent corresponding to the previous, already cancelled, mediation session. Consequently, the new mediation session would be mistakenly cancelled too. To prevent the occurrence of such a problem, a unique *id* has to be added to each mediation session. This way, a mediator could disregard obsolete **wait!** and **evaluate!** messages. The unique identification of mediation sessions is removed from the pseudo-code in order to keep it as simple as possible.

This approach may imply some kind of a live-lock, where repeatedly no agent succeeds at initiating a mediation sessions. However, such a live-lock cannot occur due to the priorities of the agents. Consider agent $x_p$ that has the highest priority among all the agents that wish to mediate. In case agent $x_p$ obtains a lock on all the agents in its *good_list*, it can initiate a mediation session and there is no live-lock. The interesting situation is when agent $x_p$ fails to get a lock on all the agents in its *good_list* (receives at least one **wait!** message). Even in this case agent $x_p$ will eventually succeed at initiating a mediation session, since all the agents in its *good_list* are aware of its willingness to mediate. The agents that are at the moment locked by other mediators (both initiated mediation sessions and mediation sessions that are to be canceled) will eventually be freed by either **cancel!** or **accept!** messages. Since these agents are all aware of agent $x_p$'s willingness to mediate, they will not join any mediation session other than agent $x_p$'s (unless $x_p$ informs them that it no longer wishes to mediate). Consequently, agent $x_p$ will eventually obtain a lock on all the agents in its *good_list* – contradicting the implied live-lock.

### 6.2 Neighboring Mediation Sessions

Sequential and concurrent neighboring mediation sessions may result in new conflicts being created without any of the *good_lists* growing. Such mediation sessions may lead to an infinite loop as depicted in section 4. $A_7$ in *configuration 2* is an example of an agent that participates in sequential neighboring mediation sessions (of the mediators $A_1$ and $A_5$). On the other hand, $A_3$ in *configuration 2* is an example of an agent whose neighbors have an incorrect view of, due to concurrent mediation sessions.

A solution to the problem of subsequent neighboring mediation sessions could be obtained if an agent (for example, $A_7$ in *configuration 2*) would agree to participate in a new mediation session only when its *agent_view* is updated with all the changes of the previous mediation session. This is achieved by the mediator sending its entire solution $s$ in the **accept!** messages, instead of just specific $d'_j$'s. Therefore, the sending of **accept!** messages (**choose_solution**, line 11) in Algorithm 2 is changed to the following:

    send (**accept!**, $(s, x_i, p_i, d_i, m_i)$) to $x_j$;

Upon receiving the revised **accept!** message (Algorithm 5), agent $i$ now updates all the $d_k$'s in the received solution $s$ accept for $d_k$'s that are not in $i$'s *agent_view* (**accept!**, lines 1-3). Notice that agent $i$ still has to send **ok?** messages to its neighbors (**accept!**, line 7), since not all of its neighbors were necessarily involved in the mediation session.

A solution to the problem of concurrent neighboring mediation sessions could be obtained if the mediator is informed *post factum* that a new conflict has been created due to concurrent mediation sessions. In this manner, the mediator can add the new conflicting agent to its *good_list*. Algorithm 5 presents the changes and additions to APO that are needed for handling concurrent neighboring mediation sessions.

When an agent $x_i$ participating in a mediation session receives the **accept!** message from its mediator, it keeps a list of all its neighbors (in the constraint graph) that are not included in the **accept!** message (not part of the mediation session), each associated with the mediator (**accept!**, lines 4-5). The list is named *conc_list*, since it contains agents that are potentially involved in concurrent mediation sessions.

Upon receiving an **ok?** message from an agent $x_j$ belonging to the *conc_list* (**ok?**, line 2), agent $x_i$ checks if the data from the received **ok?** message generates a new conflict with $x_j$ (**ok?**, line 3). If no new conflict was generated, agent $x_j$ is just removed from the *conc_list* (**ok?**, line 6). However, in case a conflict was generated (**ok?**, lines 3-5), agent $x_i$ perceives that agent $x_j$ and itself have been involved in concurrent mediation sessions that created new conflicts. In this case, agent $x_i$'s mediator should add agent $x_j$ to its *agent_view* and *good_list*. Hence, agent $x_i$ sends a new **add!** message to the mediator (associated with agent $x_j$ in the *conc_list*). When the mediator receives the **add!** message it adds agent $x_j$ to its *agent_view* and its *good_list* (**add!**, lines 1-2).

---

**Algorithm 5** Handling neighboring mediation sessions.

---
**when received** (**accept!**, $(s, x_j, p_j, d_j, m_j)$) **do**
1: **for each** $x_k \in agent\_view$ (starting with $x_i$) **do**
2:    **if** $x_k \in s$ **do**
3:       update $agent\_view$ with $(x_k, d_k)$;
4:    **else if** $d_i$ does not generate a conflict with the existing $d_k$ **do**
5:       add $(x_k, x_j)$ to $conc\_list$;
6: $mediate \leftarrow$ **false**;
7: send (**ok?**, $(x_i, p_i, d_i, m_i)$) to all $x_j \in agent\_view$;
8: update $agent\_view$ with $(x_j, p_j, d_j, m_j)$;
9: **check_agent_view**;

**when received** (**ok?**, $(x_j, p_j, d_j, m_j)$) **do**
1: update $agent\_view$ with $(x_j, p_j, d_j, m_j)$;
2: **if** $x_j \in conc\_list$ **do**
3:    **if** $d_j$ generates a conflict with $d_i$ **do**
4:       **for each** tuple $(x_j, x_k)$ in $conc\_list$ **do**
5:          send (**add!**, $(x_j)$) to $x_k$;
6:    remove all tuples $(x_j, x_k)$ from $conc\_list$;
7: **check_agent_view**;

**when received** (**add!**, $(x_j)$) **do**
1: send (**init**, $(x_i, p_i, d_i, m_i, D_i, C_i)$) to $x_j$;
2: add $x_j$ to $init\_list$;

---

There is a slight problem with this solution, since it may push the problem solving process to become centralized. This may happen because an **ok?** message from agent $x_j$ that generates a new conflict may actually have been the result of a later mediation session that agent $x_j$ was involved in. In such a case, $x_j$'s mediator already added agent $x_i$ to its *good_list*. Adding agent $x_j$ to the *good_list* of $x_i$'s mediator is not necessary for the completeness of the algorithm. It does lead to a faster convergence of the problem into a centralized one. Nevertheless, experiments show that the effect of such growth of *good_lists* is negligible (see section 9.3).

### 6.3 Preventing Busy-Waiting

To insure that partial mediation sessions do not occur, a **wait!** message received by a mediator (Algorithm 4) causes it to cancel the mediation session (section 6.1). The cancellation of the session is immediately followed by a call to **check_agent_view** (**wait!**, line 5). Such a call will most likely result in an additional attempt by the agent to start a mediation session, due to the high probability that the agent's view did not change since its previous mediation attempt. The reasons that failed the previous mediation attempt may very well cause the new mediation session not to succeed also. Such subsequent mediation attempts may occur several times before the mediation session succeeds or the mediator decides to stop its attempts. As a matter of fact, the mediator remains in a busy-waiting mode, until either its view changes, or the reasons for the mediation session's failure are no longer valid. The latter case enables the mediation session to take place.

Such a state of busy-waiting adds unnecessary overhead to the computation load of the problem solving. In particular, it increases the number of sent messages. To prevent this overhead, the mediating agent $x_m$ has to work in an interrupt-based manner rather than a busy-waiting manner. In an interrupt-based approach the mediator is notified (interrupted) when the reason for the previous mediation session's failure is no longer valid. This is done by an **ok?** message that is sent to the mediator by the agent $x_w$ that sent the preceding **wait!** message, which caused the mediation session to fail. The agent $x_w$ will send such an **ok?** message only when the reason that caused it to send the **wait!** message becomes obsolete. Namely, when one of the following occurs:

- The mediation session that $x_w$ was involved in is over.
- An agent with a higher priority than $x_m$ no longer wants to mediate.
- The *init_list* of $x_w$ has been emptied out.

In order to remember which agents have to be notified (interrupted) when one of the above instances occurs, an agent maintains a list of *pending* mediators called *wait_list*. Each time an agent sends a **wait!** message to a mediator, it adds that mediator to its *wait_list*. Whenever an agent sends **ok?** messages, it clears its *wait_list*.

A few changes to the pseudo-code must be applied in order to use the interrupt-based method. To maintain the *wait_list*, the following line has to be added to Algorithm 3 after line 3 in **evaluate?** (inside the **if** statement):

    add $x_j$ to *wait_list*;

Also, after sending **ok?** messages to the entire *agent_view*, as done for example in procedure **check_agent_view** line 7 (Algorithm 1), the following line should be added:

    empty *wait_list*;

Finally, there is need to interrupt pending mediators whenever the reason for their mediation session's failure may be no longer valid. For example, when an agent is removed from the *init_list* (**init**, line 9) in Algorithm 1, the following lines need to be added (inside the **else** statement):

    **if** *init_list* $== \emptyset$ **do**
      send (**ok?**, $(x_i, p_i, d_i, m_i)$) to all $x_w \in wait\_list$;
      empty *wait_list*;

These lines handle the case when the *init_list* has been emptied out. Similar additions must be applied to deal with the other mentioned cases. Applying this interrupt-based method rules out the need for busy-waiting. Thus, the call for **check_agent_view** (**wait!**, line 5) can be discarded.

## 7. Soundness and Completeness

In this section we will show that CompAPO is both sound and complete. Our proofs follow the basic structure and assumptions of the original APO proofs (Mailler & Lesser, 2006). The original completeness proof was incorrect because of the incompleteness of the original algorithm. Consequently, we will not use the assertion that was discussed in detail in section 3 and that played a key role in the original (and incorrect) proof of completeness (Mailler & Lesser, 2006). The following lemmas are needed for the proofs of soundness and completeness.

**Lemma 1** *Links are bidirectional. i.e. if $x_i$ has $x_j$ in its agent_view then eventually $x_j$ will have $x_i$ in its agent_view.*

**Proof (as appears in the work of Mailler and Lesser, 2006):**
Assume that $x_i$ has $x_j$ in its *agent_view* and that $x_i$ is not in the *agent_view* of $x_j$. In order for $x_i$ to have $x_j$ in its *agent_view*, $x_i$ must have received an **init** message at some point from $x_j$. There are two cases.

**Case 1:** $x_j$ is in the *init_list* of $x_i$. In this case, $x_i$ must have sent $x_j$ an **init** message first, meaning that $x_j$ received an **init** message and therefore has $x_i$ in its *agent_view* – a contradiction.

**Case 2:** $x_j$ is not in the *init_list* of $x_i$. In this case, when $x_i$ receives the **init** message from $x_j$, it responds with an **init** message. That means that if the reliable communication assumption holds, eventually $x_j$ will receive $x_i$'s **init** message and add $x_i$ to its *agent_view* – also a contradiction.

**Definition 1** *An agent is considered to be in a stable state if it is waiting for messages, but no message will ever reach it.*

**Definition 2** *A deadlock is a state in which an agent that has conflicts and desires to mediate enters a stable state.*

**Lemma 2** *A deadlock cannot occur in the CompAPO algorithm.*

**Proof:**
Assume that agent $x_i$ enters a deadlock. This means that agent $x_i$ desires to mediate, but is in a stable state. The consequence of this is that agent $x_i$ would not be able to get a lock on all the agents in its *good_list*.

One possibility is that $x_i$ already invited the members in its *good_list* to join its mediation session by sending **evaluate?** messages. After a finite time it will receive either **evaluate!** or **wait!** messages from all the agents in its *good_list*. Depending on the replies, $x_i$ either initiates a mediation session or cancels it. Either way, $x_i$ is not in a stable state – contradicting the assumption.

The other possibility is that $x_i$ did not reach the stage in which it invites other agents to join its mediation session. This can only happen, if there exists at least one agent $x_j$ that in $x_i$'s point of view both desires to mediate ($m'_j = $ **true**) and has a higher priority than $x_i$ ($p'_j > p_i$). There are two cases in which $x_j$ would not mediate a session that included $x_i$, when $x_i$ was expecting it to:

**Case 1:** $x_i$ has $m'_j =$ **true** in its *agent_view* when the actual value should be false. Assume that $x_i$ has $m'_j =$ **true** in its *agent_view* when actually $m_j =$ **false**. This would mean that at some point $x_j$ changed the value of $m_j$ to **false** without informing $x_i$ about it. There is only one place in which $x_j$ changes the value of $m_j$ – the **check_agent_view** procedure. Note that in this procedure, whenever the flag changes from **true** to **false**, the agent sends **ok?** messages to all the agents in its *agent_view*. Since by Lemma 1 we know that $x_i$ is in the *agent_view* of $x_j$, agent $x_i$ must have eventually received the message informing it that $m_j =$ **false**, contradicting the assumption.

**Case 2:** $x_j$ believes that $x_i$ should be mediating when $x_i$ does not believe it should be. In $x_j$'s point of view, $m'_i =$ **true** and $p'_i > p_j$. By the previous case, we know that if $x_j$ believes that $m_i$ is **true** ($m'_i =$ **true**) then this must be the case. We only need to show that the condition $p'_i > p_j$ is impossible. Assume that $x_j$ believes that $p'_i > p_j$ when in fact $p_i < p_j$. This means that at some point $x_i$ sent a message to $x_j$ informing it that its current priority was $p'_i$. Since we know that priorities only increase over time (all the *good_lists* can only get larger), we know that $p'_i \leq p_i$ ($x_j$ always has the correct value or an underestimate of $p_i$). Since $p_i < p_j$ and $p'_i \leq p_i$ then $p'_i < p_j$ – a contradiction to the assumption.

**Definition 3** *The algorithm is considered to be in a stable state when all the agents are in a stable state.*

**Theorem 1** *The CompAPO algorithm is sound. i.e., it reaches a stable state only if it has either found an answer or no solution exists.*

**Proof:**

We assume that all the agents reach a stable state, and consider all the cases in which this can happen.

**Case 1:** No agent has conflicts. In this case, all the agents are in a stable state and with no conflicts. This means that the current value that each agent has for its variable satisfies all its constraints. Consequently, the current values are a valid solution to the overall problem, and the CompAPO algorithm has found an answer.

**Case 2:** A no solution message has been broadcast. In this case, at least one agent found out that some subproblem has no solution, and informed all the agents about it by broadcasting a no solution message. Consequently, each agent that receives this message (all the agents) stops its run and reports that no solution exists.

**Case 3:** Some agents have conflicts. Let us consider some agent $x_i$ that has a conflict. Since it has a conflict, $x_i$ desires to mediate. If it is able to perform a mediation session then it is not in a stable state in contradiction to the assumption. Therefore, the only condition in which $x_i$ can remain in a stable state is if it is expecting a mediation request from a higher priority agent $x_j$ that does not send it – in other words, when it is deadlocked. By Lemma 2 this cannot happen.

Since only cases 1 and 2 can occur, the CompAPO algorithm reaches a stable state only if it has either found an answer or no solution exists. Consequently, the CompAPO algorithm is sound. □

**Lemma 3** *If there exist agents that hold the entire graph in their good_list and desire to mediate, then one of these agents will perform a mediation session.*

**Proof:**

We shall consider two cases – when there is only one such agent that holds the entire graph in its *good_list* and desires to mediate, and when there are several such agents.

**Case 1:** Consider agent $x_i$ to be the only agent that holds the entire graph in its *good_list* and desires to mediate. Since $x_i$ has the entire graph in its *good_list* it has the highest possible priority. Moreover, all of the agents are aware of $x_i$'s priority ($p_i$) and desire to mediate ($m_i$) due to **ok?** messages they received from $x_i$ containing this information ($x_i$ sent **ok?** messages to all the agents in its *agent_view*, which holds the entire graph). Consequently, no agent will engage from this point on, in any mediation session other than $x_i$'s. Since all mediation sessions are finite and no new mediation sessions will occur, agent $x_i$ will eventually get a lock on all the agents and will perform a mediation session.

**Case 2:** If several such agents exist, then the tie in the priorities is broken by the agents' index. Consider $x_i$ to be the one with the highest index out of these agents, and apply the same proof of case 1.

**Lemma 4** *If an agent holding the entire graph in its good_list performs a mediation session, the algorithm reaches a stable state.*

**Proof:**

Consider the mediator to be agent $x_i$. Following the first part of CompAPO's solution (section 6.1), an agent can perform a mediation session only if it received **evaluate!** messages from all the agents in its *good_list*. Since $x_i$ holds the entire graph in its *good_list*, it means that all the agents in the graph have sent **evaluate!** messages to $x_i$ and set their mediate flags to be **true**. This means that until $x_i$ completes its search and returns **accept!** messages with its solution, no agent can change its assignment. Assuming that the centralized internal solver that $x_i$ uses is sound and complete, it will find a solution to the entire problem if such a solution exists, or alternatively conclude that no solution exists. If no solution exists, then $x_i$ informs all the agents about this and the problem solving terminates. Otherwise, each agent receives the **accept!** message from $x_i$ that contains the solution to the entire problem. Consequently, no agent has any conflicts and the algorithm reaches a stable state.

**Lemma 5** *Infinite value changes without any mediation sessions cannot occur.*

**Proof:**

The proof will focus on line 6 of the **check_agent_view** procedure, the only place in the code in which a value is changed without a mediation session. As a reminder, notice that all the agents in the graph are ordered by their priority (ties are broken by the IDs of the agents).

Consider the agent with the lowest priority ($x_{p_1}$). Agent $x_{p_1}$ cannot change its own value, since line 5 in the **check_agent_view** procedure states that in order to reach the value change in line 6, the current value must conflict exclusively with lower priority agents. Clearly this is impossible for agent $x_{p_1}$, which has the lowest priority in the graph.

Now, consider the next agent in the ordering, $x_{p_2}$. Agent $x_{p_2}$ can change its current value when it is in conflict exclusively with lower priority agents. The only lower priority agent in this case is $x_{p_1}$. If $x_{p_1}$ and $x_{p_2}$ are not neighbors, then agent $x_{p_2}$ cannot change its own value for the same reason as agent $x_{p_1}$. Otherwise, agent $x_{p_2}$ will know the up-to-date value of agent $x_{p_1}$ in finite time (any previously sent updates regarding $x_{p_1}$'s value will eventually reach agent $x_{p_2}$), since we proved that the value of $x_{p_1}$ cannot be changed without a mediation session. After $x_{p_2}$ has the up-to-date value of all its lower priority neighbors (only $x_{p_1}$), it can change its own value at most once without a mediation session. Eventually, all the neighbors of $x_{p_2}$ will be updated with the final change of its value.

In general, any agent $x_{p_i}$ (including the highest priority agent) will in finite time have the up-to-date values of all its lower priority agents. When this happens, it can change its own value at most once without a mediation session. Eventually, all the neighbors of $x_{p_i}$ will be updated with the final change of its value. Thus, infinite value changes without any mediation sessions cannot occur.

This proof implicitly relies on the fact that the ordering of agents does not change. However, the priority of an agent may change in time. Nevertheless, the priorities are bounded by the size of the graph, so the number of priority changes is finite. This proves that value changes cannot indefinitely occur without any mediation sessions. ∎

**Lemma 6** *If from this point on no agent will mediate or desire to mediate, the algorithm will reach a stable state.*

**Proof:**

We shall consider two cases – when there are no messages that have not yet arrived to their destinations, and when there are such messages.

**Case 1:** Consider the case when there are no messages that have not yet arrived to their destination. If no agent desires to mediate, then all the $m_i$'s are **false**, meaning that no agent in the graph has conflicts. Consequently, the current state of the graph is a solution that satisfies all the constraints, and the algorithm reaches a stable state.

**Case 2:** Consider the case when there are some messages that have not yet arrived to their destinations. Eventually these messages will arrive. According to the assumption of the lemma, the arrival of these messages will not make any of the agents desire to mediate. Next, we consider the arrival of each type of message and show that it cannot lead to infinite exchange of messages:

- **evaluate?**, **evaluate!**, **wait!**, **cancel!**: These messages must belong to an obsolete mediation session, or otherwise contradict the assumption of the lemma. Accordingly, they may result in some limited exchange of messages (e.g., sending **wait!** in line 3 of **evaluate?**). Some of these messages may lead to a call to the **check_agent_view** procedure.
- **accept!**: This message cannot be received without contradicting the assumption, since the receiving agent has to be in an active mediation session when receiving an **accept!** message.
- **init**: This message is part of a handshake between two agents. Consequently, at most a single additional **init** message will be sent. This leads to a call to the **check_agent_view** procedure by each of the involved agents.

- **add!**: This message results in the sending of a single **init** message.
- **ok?**: This message may result in the sending of a finite number of **add!** messages. It also leads to a call to the **check_agent_view** procedure.

By examining all the types of messages, we conclude that each message can at most lead to a finite exchange of messages, and to a finite number of calls to the **check_agent_view** procedure. We only need to show that a call to **check_agent_view** cannot lead to infinite exchange of messages.

The **check_agent_view** procedure has 4 possible outcomes. It may simply return (line 2), change the value of the variable (lines 6-7), mediate (line 9), or update its desire to mediate (lines 11-12). According to the assumption of the lemma, it cannot mediate. An update of its desire to mediate, means that the value of $m_i$ was true, or will be updated to true. Either way, this is again in contradiction to the assumption of the lemma. Consequently, the only possibilities are a simple return, or a change in the value of its own variable. According to Lemma 5, such value changes cannot indefinitely occur without any mediation sessions. Consequently, the final messages will eventually arrive to their destinations, and the first case of the proof will hold.

**Definition 4** *One says that the algorithm advances if at least one of the good_lists grows.*

**Lemma 7** *After every n mediation sessions, the algorithm either advances or reaches a stable state.*

**Proof:**

Consider a mediation session of agent $x_i$. The mediation session has three possible outcomes – no solution satisfying the constraints within the *good_list*, a solution satisfying the constraints within the *good_list* but with violations of external constraints, and a solution satisfying all the constraints within the *good_list* and all the external constraints.

**Case 1:** No solution that satisfies the constraints within $x_i$'s *good_list* exists, therefore the entire problem is unsatisfiable. In this case, $x_i$ informs all the agents about this and the problem solving terminates.

**Case 2:** $x_i$ finds a solution that satisfies the constraints within its *good_list* but violates external constraints. In this case, $x_i$ adds the agents with whom there are external conflicts to its *good_list*. These agents were not already in $x_i$'s *good_list*, since the mediation session included the entire *good_list* of $x_i$ (according to section 6.1). Consequently, $x_i$'s *good_list* grows and the algorithm advances.

**Case 3:** $x_i$ finds a solution that satisfies the constraints within its *good_list* and all the external constraints. Following the second part of CompAPO's solution (section 6.2), agents from $x_i$'s *good_list* maintain a *conc_list*, and would notify $x_i$ to add agents to its *good_list* in case they experience new conflicts due to concurrent mediation sessions. In such a case, $x_i$ would be notified, its *good_list* would grow and the algorithm would advance.

The only situation in which the algorithm does not advance or reach a stable state, is when all the mediation sessions experience case 3, and no concurrent mediation sessions create new conflicts. In that case, after at most $n$ mediation sessions (equal to the overall number of agents), all the agents would have no desire to mediate. According to Lemma 6, the algorithm reaches a stable state.

**Lemma 8** *If there exists a group of agents that desire to mediate, a mediation session will eventually occur.*

**Proof:**

No agent will manage to get a lock on all the agents in its *good_list* (essential for a mediation session to occur) only if all the agents in the group that sent **evaluate?** messages got at least one **wait!** message each. If this is the case, consider $x_i$ to be the highest priority agent among this group.

Each **wait!** message that agent $x_i$ received is either from an agent that is a member of the group or from an agent outside the group, currently involved in another mediation session. In case this agent ($x_j$) belongs to the group, $x_j$ also got some **wait!** message (clearly, this *wait!* message arrived after $x_j$ sent *wait!* to $x_i$). $x_j$ will therefore cancel its mediation session, and will wait for $x_i$'s next **evaluate?** message (since $x_j$ is now aware of $x_i$'s desire to mediate and $p_i$ is the highest priority among the agents that currently desire to mediate). In case $x_j$ does not belong to the group, the mediation session that $x_j$ is involved in will eventually terminate, and $x_i$ will get the lock, unless $x_j$ has a higher priority than $x_i$ ($p_j > p_i$) and also $x_j$ desires to mediate when the session terminates. If this is the case, $x_j$ will eventually get the lock for the same reasons.

**Theorem 2** *The CompAPO algorithm is complete. i.e., if a solution exists, the algorithm will find it, and if a solution does not exist, the algorithm will report that fact.*

**Proof:**

Since we have shown in Theorem 1 that whenever the algorithm reaches a stable state, the problem is solved and that when it finds a subset of variables that is unsatisfiable it terminates, we only need to show that it always reaches one of these two states in a finite time.

According to Lemma 6, if from some point in time no agent will mediate or desire to mediate, the algorithm will reach a stable state. According to Lemma 8 if there exist agents that desire to mediate, eventually a mediation session will occur. From Lemmas 6 and 8 we conclude that the only possibility for the algorithm not to reach a stable state is by continuous occurrences of mediation sessions. According to Lemma 7, after every n mediation sessions, the algorithm either advances or reaches a stable state. Consequently, the algorithm either reaches a stable state or continuously advances.

In case the algorithm continuously advances, the *good_lists* continuously grow. At some point, some agents (eventually all the agents) will hold the entire graph in their *good_list*. One of these agents will eventually desire to mediate (if not, then according to Lemma 6, the algorithm reaches a stable state). According to Lemma 3, one of these agents will perform a mediation session. According to Lemma 4, the algorithm reaches a stable state. □

## 8. OptAPO – an Optimizing APO

Distributed Constraint Optimization Problems (DisCOPs) are a version of distributed constraint problems, in which the goal is to find an optimal solution to the problem, rather than a satisfying one. In an optimization problem, an agent associates a cost with violated constraints and maintains bounds on these costs in order to reach an optimal solution that minimizes the number of violated constraints.

A number of algorithms were proposed in the last few years for solving DisCOPs. The simplest algorithm of these is Synchronous Branch and Bound (SyncBB) (Hirayama & Yokoo, 1997), which is a distributed version of the well-known centralized Branch and Bound algorithm. Another algorithm which uses a Branch and Bound scheme is Asynchronous Forward Bounding (AFB) (Gershman, Meisels, & Zivan, 2006), in which agents perform sequential assignments which are propagated for bounds checking and early detection of a need to backtrack. A number of algorithms use a pseudo-tree which is derived from the structure of the DisCOP in order to improve the process of acquiring a solution for the optimization problem. ADOPT (Modi, Shen, Tambe, & Yokoo, 2005) is such an asynchronous algorithm in which assignments are passed down the pseudo-tree. Agents compute upper and lower bounds for possible assignments and send costs up to their parents in the pseudo-tree. These costs are eventually accumulated by the root agent. Another algorithm which exploits a pseudo tree is DPOP (Petcu & Faltings, 2005). In DPOP, each agent receives from the agents which are its sons in the pseudo-tree, all the combinations of partial solutions in their sub-tree and their corresponding costs. The agent calculates and generates all the possible partial solutions which include the partial solutions it received from its sons and its own assignments and sends the resulting combinations up the pseudo-tree. Once the root agent receives all the information from its sons, it produces the optimal solution and propagates it down the pseudo-tree to the rest of the agents.

Another very different approach was implemented in the Optimal Asynchronous Partial Mediation (OptAPO) (Mailler & Lesser, 2004; Mailler, 2004) algorithm, which is an optimization version of the APO algorithm. Differently to APO, the OptAPO algorithm introduces a second type of mediation sessions called *passive* mediation sessions. The goal of the passive sessions is to update the bounds on the costs without changing the values of variables. These sessions add parallelism to the algorithm and accelerate the distribution of information. This might solve many problems that result from incorrect information, which is discussed in section 5.2. However, *active* mediation sessions also occur in OptAPO. The active sessions may consist of parts of the *good_list* (partial mediation sessions), and as a result lead to the problems described in section 5.1. Moreover, a satisfiable problem should also be solved by OptAPO, returning a zero optimal cost. Therefore, the infinite loop scenario described in section 4 will also occur in OptAPO, which behaves like APO when the problem is satisfiable.

The OptAPO algorithm must be corrected in order for the aforementioned problems to be solved. In section 6 several modifications to the APO algorithm are proposed. These changes turn APO into a complete search algorithm – CompAPO. Equivalent modifications must also be applied to the OptAPO algorithm in order to ensure its correctness. Interestingly, these modifications to APO are to procedures that are similar in APO and OptAPO. The main differences between APO and OptAPO are in the addition of passive mediation sessions (procedure **check_agent_view**) to OptAPO, and in the internal search that mediators perform (procedure **choose_solution**). However, neither of these procedures is effected by the modifications of CompAPO. Thus, the pseudo-code of the changes that must be applied to OptAPO is very similar to the modifications of CompAPO, and is therefore omitted from this paper. The performance of the resulting algorithm – CompOptAPO – is evaluated in section 9.5. The full pseudo-code of the original OptAPO algorithm can be found in the work of Mailler and Lesser (2004).

## 9. Experimental Evaluation

The original (and incomplete) version of the APO algorithm was evaluated by Mailler and Lesser (2006). It was compared to the AWC algorithm (Yokoo, 1995), which is not an efficient DisCSP solver (Zivan et al., 2007). The experiments were performed on 3-coloring problems, which are a subclass of uniform random constraints problems. These problems are characterized by a small domain size, low constraints density, and fixed constraints tightness (for the characterization of random CSPs see the works of Prosser, 1996 and Smith, 1996). The comparison between APO and AWC (Mailler & Lesser, 2006) was made with respect to three measures – the number of *sent messages*, the number of *cycles*, and the *serial runtime*. While the number of sent messages is a very important and widely accepted measure, the other measures are problematic. During a cycle, incoming messages are delivered, the agent is allowed to process the information, and any messages that were created during the processing are added to the outgoing queue to be delivered at the beginning of the next cycle. The meaning of such a cycle in APO is that a mediation session that possibly involves the entire graph takes just a single cycle. Such a measure is clearly problematic, since every centralized algorithm solves a problem in just one cycle. Measuring the serial runtime is also not adequate for distributed CSPs, since it does not take into account any concurrent computations during the problem solving. In order to measure the concurrent runtime of DisCSP algorithms in an implementation independent way, one needs to count non-concurrent constraint checks (NCCCs) (Meisels, Razgon, Kaplansky, & Zivan, 2002). This measure has gained global agreement in the DisCSP and DisCOP community (Bessiere et al., 2005; Zivan & Meisels, 2006b) and will be used in the present evaluation.

The modifications of CompAPO, and especially the prevention of partial mediation sessions (section 6.1) add synchronization to the algorithm, which may tax heavily the performance of the algorithm. Thus, it is important to evaluate the effect of these changes by comparing CompAPO to other (incomplete) versions of the APO algorithm. Additionally, we evaluate the effectiveness of the interrupt-based method compared to busy-waiting.

### 9.1 Experimental Setup

In all our experiments we use a simulator in which agents are simulated by threads, which do not hold any shared memory and communicate only through message passing. The network of constraints, in each of our experiments, is generated randomly by selecting the probability $p_1$ of a constraint among any pair of variables and the probability $p_2$, for the occurrence of a violation among two assignments of values to a constrained pair of variables. Such uniform random constraints networks of $n$ variables, $k$ values in each domain, a constraints *density* of $p_1$ and *tightness* $p_2$ are commonly used in experimental evaluations of CSP algorithms (Prosser, 1996; Smith, 1996).

Experiments were conducted for several density values. Our setup included problems generated with 15 agents ($n = 15$) and 10 values ($k = 10$). We drew 100 different instances for each combination of $p_1$ and $p_2$. Through all our experiments each agent holds a single variable.

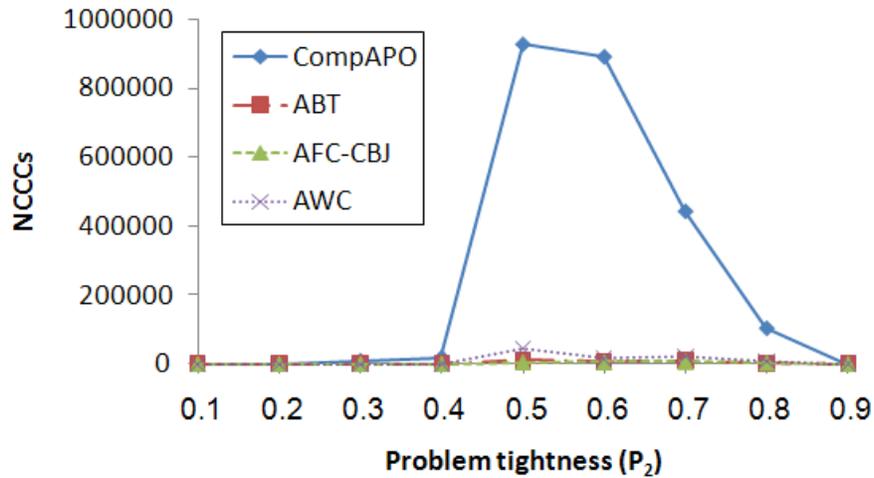

Figure 3: Mean NCCCs in sparse problems ($p_1 = 0.1$).

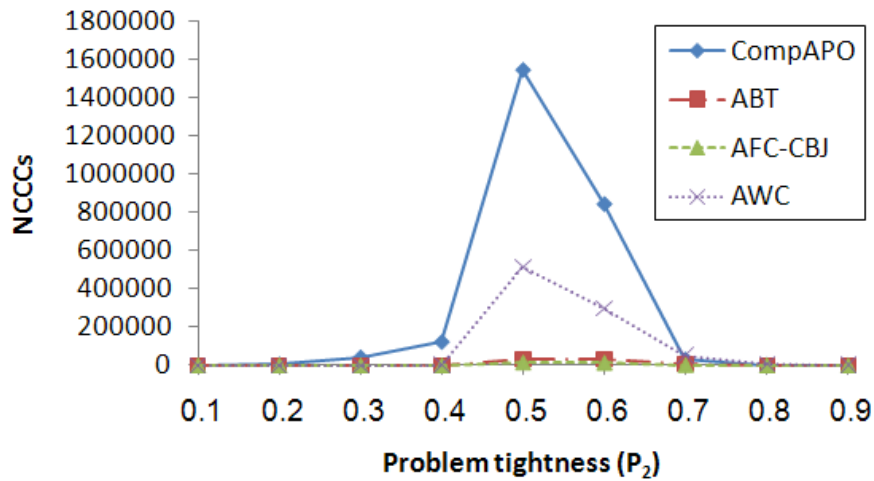

Figure 4: Mean NCCCs in medium density problems ($p_1 = 0.4$).

## 9.2 Comparison to Other Algorithms

The performance of CompAPO is compared to three asynchronous search algorithms – the well known Asynchronous Backtracking (ABT) (Yokoo et al., 1998; Yokoo & Hirayama, 2000), the extremely efficient Asynchronous Forward-Checking with Backjumping (AFC-CBJ) (Meisels & Zivan, 2007), and to Asynchronous Weak Commitment (AWC) (Yokoo, 1995), which was used in the original APO evaluation (Mailler & Lesser, 2006).

Results are presented for three sets of tests with different values of problem density – sparse ($p1 = 0.1$), medium ($p1 = 0.4$), and dense ($p1 = 0.7$). In all the sets the value of $p_2$ varies between 0.1 and 0.9, to cover all ranges of problem difficulty.

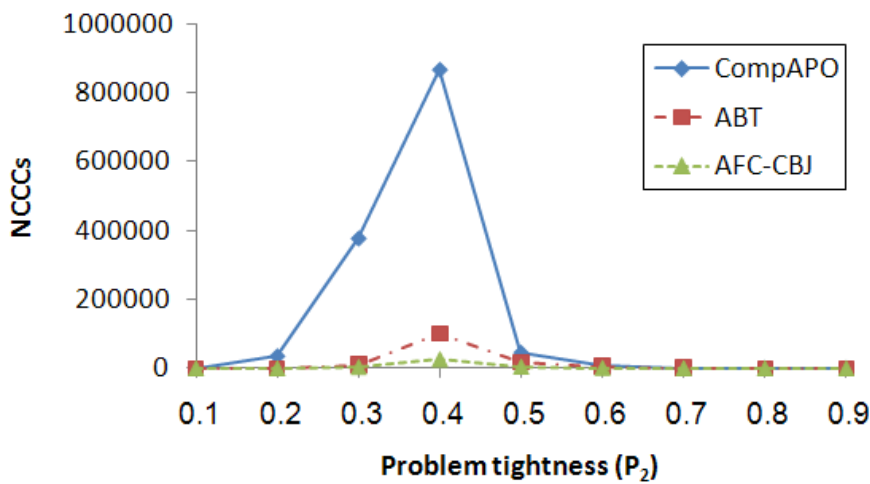

Figure 5: Mean NCCCs in dense problems ($p_1 = 0.7$).

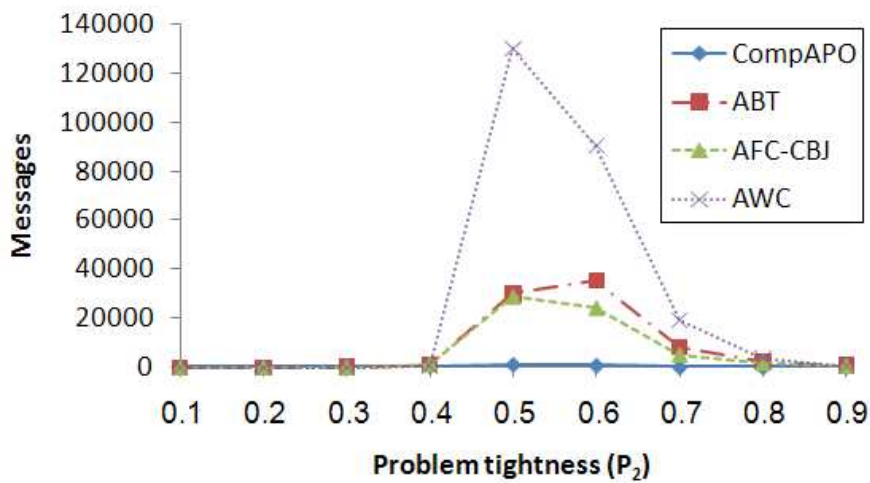

Figure 6: Mean number of messages in medium density problems ($p_1 = 0.4$).

In order to evaluate the performance of the algorithms, two independent measures of performance are used – search effort in the form of NCCCs and communication load in the form of the total number of messages sent. Figures 3, 4, and 5 present the number of NCCCs performed by CompAPO while solving problems with different densities. Figure 6 shows the total number of messages sent during the problem solving process. All figures exhibit the phase-transition phenomenon – for increasing values of the tightness, $p_2$, problem difficulty increases, reaches a maximum, and then drops back to a low value. This is termed the easy-hard-easy transition of hard problems (Prosser, 1996), and was observed for DisCSPs (Meisels & Zivan, 2007; Bessiere et al., 2005).

The performance of CompAPO in NCCCs turns out to be very poor in the phase transition region compared to other asynchronous search algorithms. The worst results are when the problems are relatively sparse (Figures 3 and 4). However, even for dense problems both ABT and AFC-CBJ clearly outperform CompAPO (Figure 5). When comparing CompAPO to AWC, the results are significantly different. AWC is known to perform best in sparse problems. Thus, like ABT and AFC-CBJ it clearly outperforms CompAPO for such problems (Figure 3). For medium density problems, AWC still performs better than CompAPO but the difference between the performances of these algorithms is much smaller (Figure 4). For dense problems, AWC performs extremely bad with about ten times more NCCCs than CompAPO. The results of AWC are omitted from Figure 5, since it did not finish running this set of tests in a reasonable time and we had to stop its run after 40 hours.

Notice that the scale in Figure 4 is different than in Figures 3 and 5. This is due to especially poor performance of APO around the phase transition of medium density problems. Such behavior is untypical, since most DisCSP algorithm suffer from their worst performance around the phase transition of high density problems (Figure 5). The fact that the performance of CompAPO is better on high density problems than on medium density ones can be explained by the faster convergence to a centralized solution in dense problems. In problems around the phase transition, the CompAPO algorithm frequently reaches full centralization anyway. Thus, the faster convergence to a centralized solution actually improves the performance of the algorithm.

While the search effort performed by the agents running CompAPO is extremely high, the communication load on the system remains particularly low. This can be seen in Figure 6, for medium density problems. Similar results were achieved for sparse and dense problems. This is not surprising, since the major part of the search effort is carried out by agents performing mediation sessions without the need for an extensive exchange of messages.

### 9.3 Comparison to Other Versions of APO

Several versions of the APO algorithm were proposed by Benisch and Sadeh (2006). One of these versions (APO-BT) uses simple backtracking as its mediation procedure, instead of the Branch and Bound that was originally proposed with APO (APO-BB). The performance of CompAPO is compared to these two incomplete versions of the algorithm.

The modifications of CompAPO, and especially the prevention of partial mediation sessions (section 6.1) add synchronization to the algorithm. A potential partial mediation session must wait for other sessions to end until the mediator is able to get a lock on its entire *good_list*. Such synchronization may tax the performance of the algorithm. Nevertheless, our experiments show that CompAPO actually performs slightly better than APO-BB as measured by NCCCs (Figures 7 and 8). The improved performance can be explained by the better distribution of data when the entire solution is sent with the **accept!** message (section 6.2). Figure 9 shows that the effect of CompAPO's modifications is even greater on the communication load. This substantial advantage of CompAPO may be explained by the use of the interrupt-based approach (section 6.3) that helps performance by eliminating the unnecessary overhead of busy-waiting.

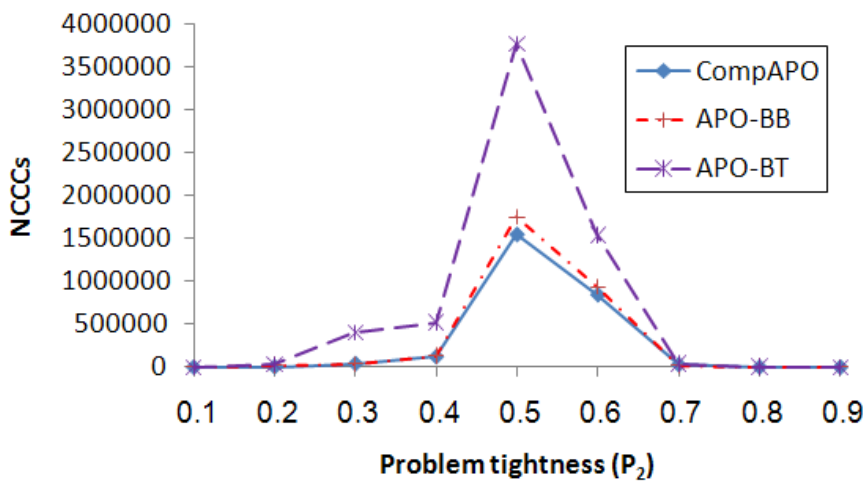

Figure 7: Mean NCCCs in medium density problems ($p_1 = 0.4$).

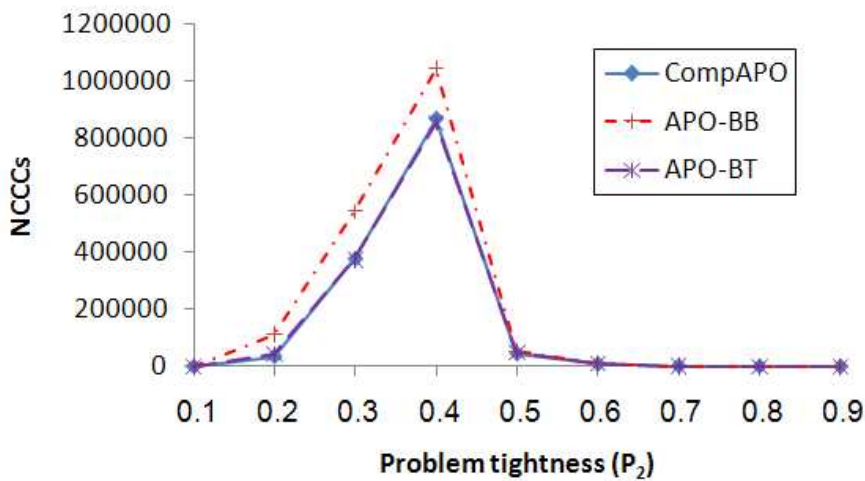

Figure 8: Mean NCCCs in dense problems ($p_1 = 0.7$).

Figure 10 presents the mean size of the largest mediation session occurring during search, for medium density problems ($p1 = 0.4$) with 15 variables. The average size of the largest mediation session is around 12 (out of a maximum of 15). It occurs for problems in the phase transition region when $p_2$ is 0.5 and 0.6. Although this number is not very far from the maximum of 15, it does suggest that a considerable portion of the hard problems are solved without reaching a full centralization.

The part of the code of CompAPO that solves the neighboring mediation sessions problem (section 6.2) implies a potential additional growth to *good_lists* (**ok?**, line 5), which may result in a faster centralization during problem solving. Nevertheless, Figure 10 clearly shows that CompAPO does not centralize faster than the original version of APO (APO-BB), except for very tight, unsolvable problems.

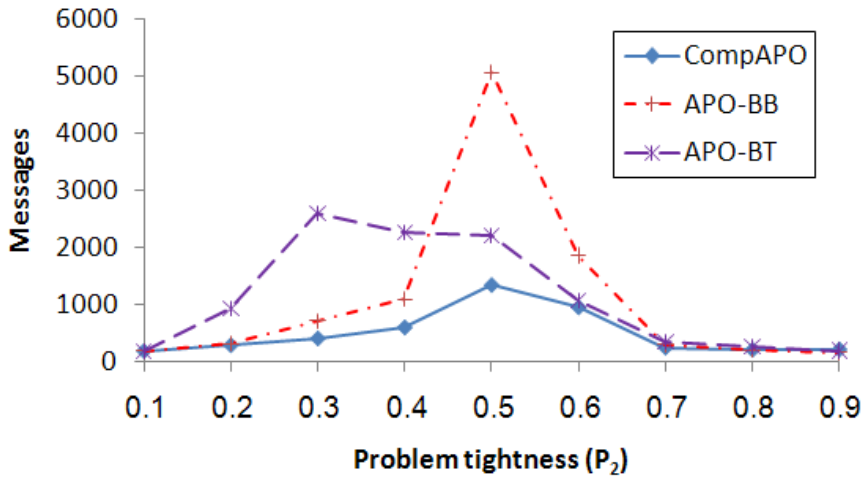

Figure 9: Mean number of messages in medium density problems ($p_1 = 0.4$).

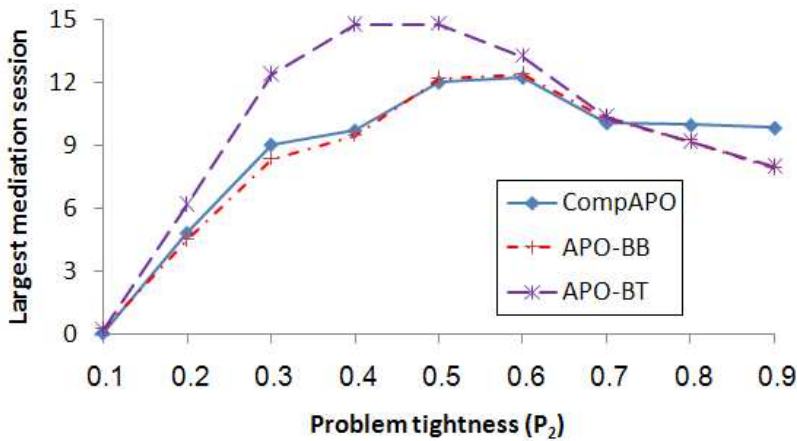

Figure 10: Mean size of the largest mediation session ($p_1 = 0.4$ and $n = 15$).

Our experiments show that in medium density problems, the APO-BT version performs poorly with respect to both NCCCs and the number of sent messages in comparison to APO-BB and CompAPO (Figures 7 and 9). The reason for ABO-BT's poor performance can be easily explained by its frequent convergence to full centralization as shown in Figure 10. Nevertheless, APO-BT has a lower communication load than APO-BB in the phase transition. The reason for this is actually the same reason that leads to APO-BT's extensive search effort. A prompt convergence to full centralization yields a high search effort (NCCCs), but at the same time may reduce the communication load.

Figure 8 shows that on dense problems APO-BT performs better than APO-BB and almost the same as CompAPO. This supports the results reported by Benisch and Sadeh (2006) for dense random DisCSPs. The same paper also presents the results for structured 3-coloring problems, in which APO-BT is outperformed by APO-BB. Similar behavior is observed in the experiments that we conducted on sparser problems (Figure 7), which suggests that the variance in APO-BT's performance has more to do with the density of the problem than with its structure.

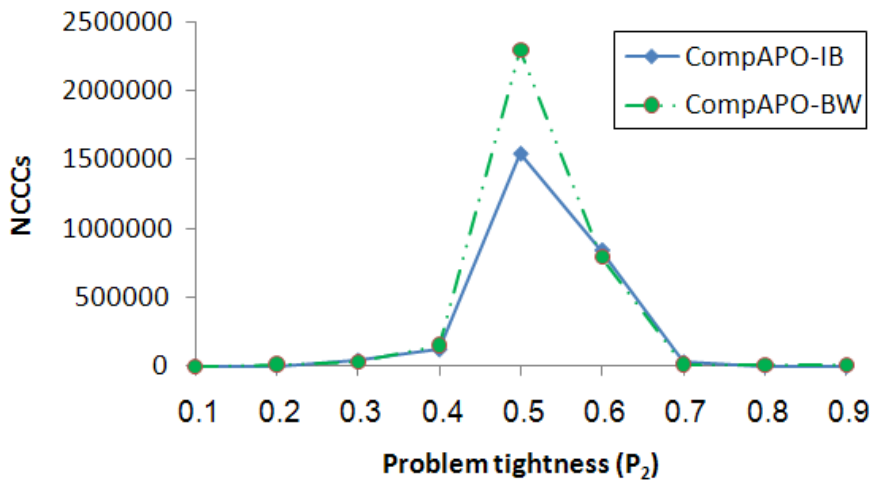

Figure 11: Interrupt-based vs. busy-waiting (mean NCCCs with $p_1 = 0.4$).

Benisch and Sadeh propose an additional version to the APO algorithm, in which the mediation session selection rule is the inverse of the original selection rule (Benisch & Sadeh, 2006). The version called IAPO instructs agents to choose the smallest mediation session rather than the largest one. It is not clear that IAPO can be turned into a correct algorithm, since the correctness proofs presented in section 7 rely on the fact that the largest mediation sessions are chosen. Consequently, the evaluation of IAPO is omitted from this paper.

### 9.4 Interrupt-Based Versus Busy-Waiting

Figures 11 and 12 present two measures of performance comparing different methods for synchronization that is needed in order to avoid conflicts between concurrent mediation sessions – interrupt-based and busy-waiting (section 6.3). The interrupt-based method clearly outperforms busy-waiting for harder problem instances. Predictably, the difference in performance is more pronounced when measuring the number of messages (Figure 12).

### 9.5 Evaluation of CompOptAPO

The original (and incomplete) version of the OptAPO algorithm was evaluated by Mailler and Lesser (2004). It was compared to the ADOPT algorithm (Modi et al., 2005), which is not the best DisCOP solver. Similarly to the original results of APO (Mailler & Lesser, 2006), the comparison between OptAPO and ADOPT (Mailler & Lesser, 2004) was made with respect to three measures – the number of *sent messages*, the number of *cycles*, and the *serial runtime*. For the same reasons as in DisCSP algorithms, cycles and serial runtime are also problematic for measuring the performance of DisCOP algorithms. As was the case with CompAPO, the CompOptAPO algorithm will also be evaluated by counting NCCCs and the number of sent messages.

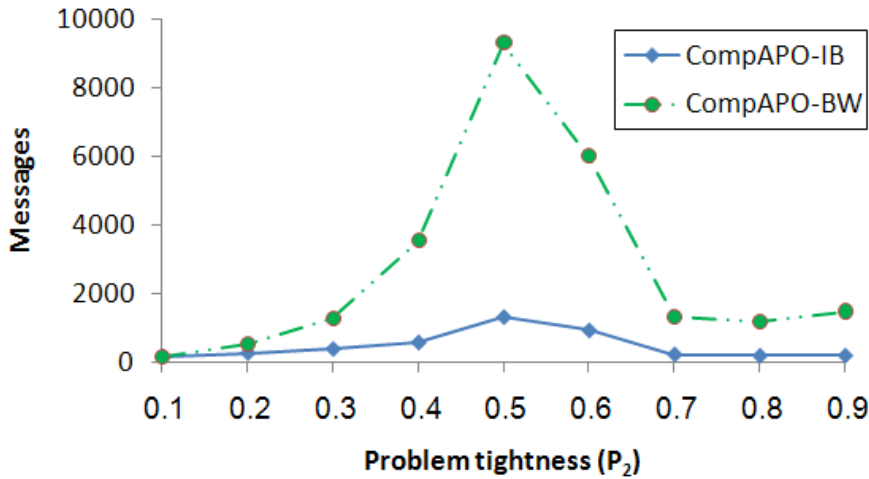

Figure 12: Interrupt-based vs. busy-waiting (mean number of messages with $p_1 = 0.4$).

The Distributed Optimization problems used in the following experiments are random Max-DisCSPs. Max-DisCSP is a subclass of DisCOP in which all constraint costs (weights) are equal to one (Modi et al., 2005). This feature simplifies the task of generating random problems, since by using Max-DisCSPs one does not have to decide on the costs of the constraints. Max-CSPs are commonly used in experimental evaluations of constraint optimization problems (COPs) (Larrosa & Schiex, 2004). Other experimental evaluations of DisCOPs include graph coloring problems (Modi et al., 2005; Zhang, Xing, Wang, & Wittenburg, 2005), which are a subclass of Max-DisCSP. The advantage of using random Max-DisCSP problems is the fact that they create an evaluation framework that is known to exhibit the phase-transition phenomenon in centralized COPs. This is important when evaluating algorithms for solving DisCOPs, enabling a known analogy with behavior of centralized algorithms as the problem difficulty changes. The problems solved in this section are randomly generated Max-DisCSP with 10 agents ($n = 10$) and 10 values ($k = 10$), constraint density of either $p_1 = 0.4$ or $p_1 = 0.7$, and varying constraint tightness $0.4 \leq p_2 < 1$.

The performance of CompOptAPO is compared to three search algorithms – Synchronous Branch and Bound (SyncBB) (Hirayama & Yokoo, 1997), AFB (Gershman et al., 2006), and ADOPT (Modi et al., 2005). ADOPT was used in the original OptAPO evaluation (Mailler & Lesser, 2004).

It must be noted that in our experiments with the original OptAPO algorithm, we have experienced several runs in which the algorithm failed to advance and did not reach a solution. This shows that the termination problem of OptAPO occurs in practice and not just in theory, for a scenario that involves particular message delays as the one presented in section 4 for the APO algorithm. Additionally, we discovered that OptAPO may not always be able to report the optimal cost (i.e., the number of broken constraints in our *Max-DisCSP* experiments). To understand how this can happen consider an "almost" disjoint graph, such as the one depicted in Figure 13. In this example we assume that agents $A_1$ and $A_2$ do not have any conflicts. Consequently, the knowledge regarding the cost $F$ is not exchanged between the two groups, and no agent holds the correct overall cost of the problem ($F = 3$). Nevertheless, when the OptAPO algorithm terminates, it does so with the optimal solution. Thus, the optimal value can be derived upon termination by summing the number of broken constraints of all of the agents. The result must be divided by two to account for broken constraints being counted by each of the involved agents.

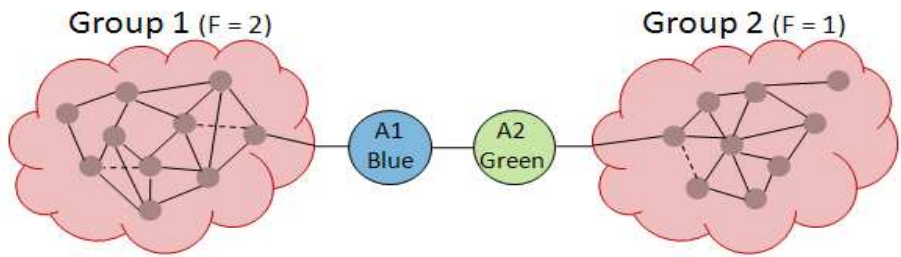

Figure 13: An example 3-coloring problem.

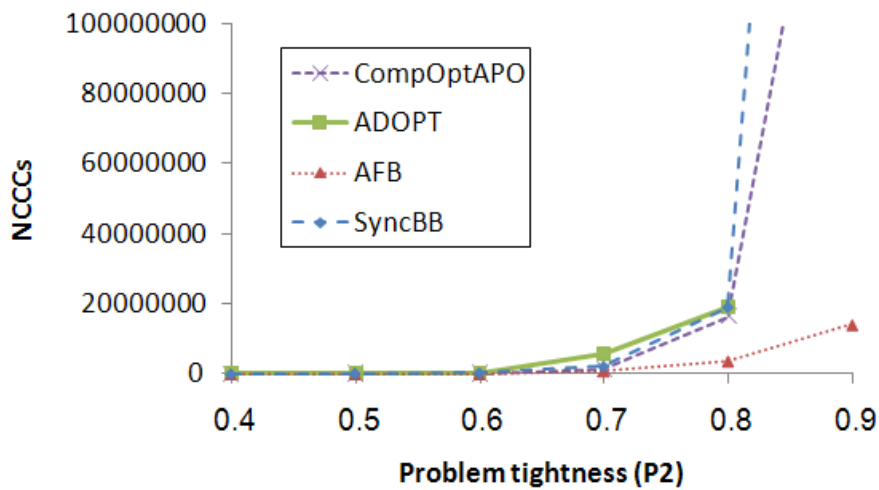

Figure 14: Mean NCCCs in sparse optimization problems ($p_1 = 0.4$).

The performance of CompOptAPO in NCCCs is comparable with other DisCOP algorithms when the problems are relatively loose (low $p_2$ value), with only the ADOPT algorithm performing slightly better. This is the case for both sparse and dense problems (Figures 14 and 15, respectively). As the problems become tighter, CompOptAPO clearly outperforms both ADOPT and SyncBB. In fact, the ADOPT algorithm failed to terminate in reasonable time for tight problems ($p_2 > 0.8$ in Figure 14 and $p_2 > 0.6$ in Figure 15). However, on tight problems the AFB algorithm is much faster than CompOptAPO. Actually, AFB was the only algorithm in our experiments that managed to terminate in reasonable time for problems that are both dense ($p_1 = 0.7$) and tight ($p_2 = 0.9$).

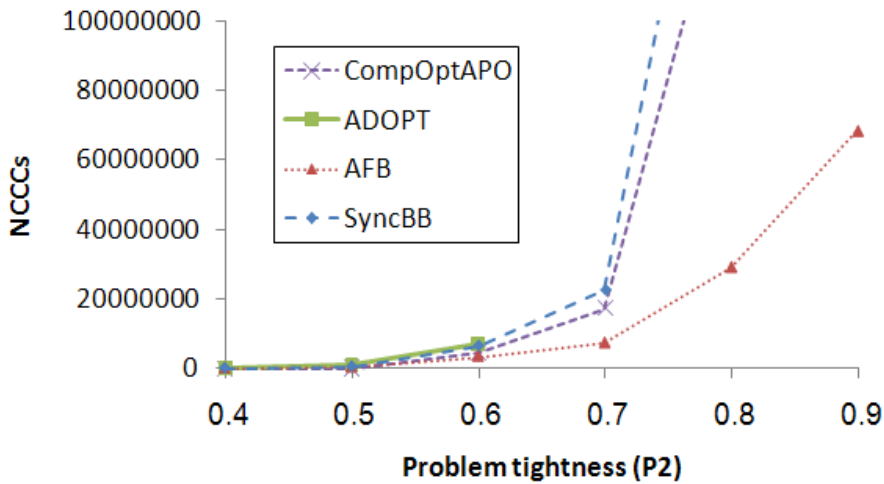

Figure 15: Mean NCCCs in dense optimization problems ($p_1 = 0.7$).

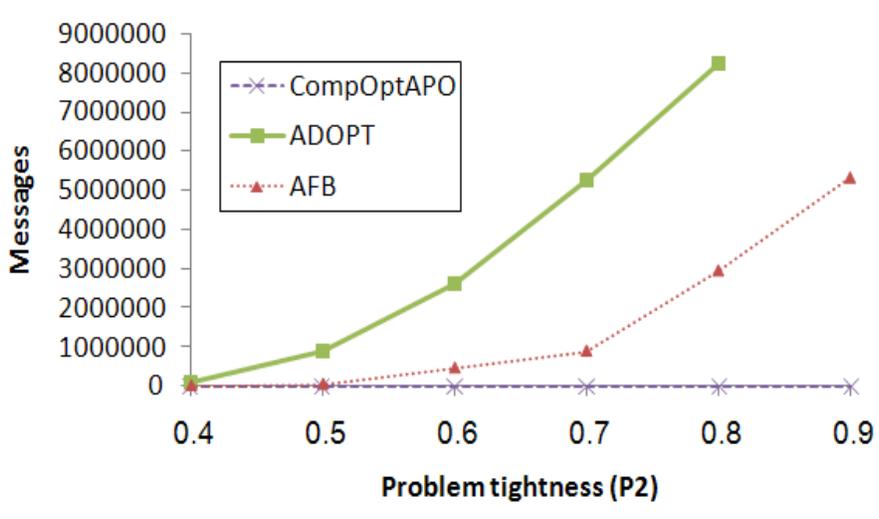

Figure 16: Mean number of messages in dense optimization problems ($p_1 = 0.7$).

Similarly to CompAPO, the communication load on the system remains particularly low when running the CompOptAPO algorithm. This can be seen in Figure 16 for dense problems. Similar results are observed for sparse problems. This is not surprising, since the major part of the search effort is carried out by agents performing mediation sessions without the need for an extensive exchange of messages.

## 10. Conclusions

The APO search algorithm is an asynchronous distributed algorithm for DisCSPs. The algorithm partitions the search into different subproblems. Each subproblem is solved by a selected agent – the mediator. When conflicts arise between a solution to a subproblem and its neighboring agents, the conflicting agents are added to the subproblem. Ideally, the algorithm either leads to compatible solutions of constraining subproblems, or to the growth of subproblems whose solution is incompatible with neighboring agents. This two-option situation was used in the original APO paper (Mailler & Lesser, 2006) to prove the termination and completeness of the algorithm.

The proof of completeness of the APO algorithm as presented by Mailler and Lesser (2006) is based on the growth of the size of the subproblems. It turns out that this expected growth of groups does not occur in some situations, leading to a termination problem of the algorithm. The present paper demonstrates this problem by following an example that does not terminate. Furthermore, the paper identifies the problematic parts in the original algorithm that interfere with its completeness and applies modifications that solve the problematic parts. The resulting CompAPO algorithm ensures the completeness of the search. Formal proofs for the soundness and completeness of CompAPO are presented.

The CompAPO algorithm forms a class by itself of DisCSP search algorithms. In contrast to backtracking or concurrent search processes, it achieves concurrency by solving subproblems concurrently. It is therefore both interesting and important to evaluate the performance of CompAPO and to compare it to other DisCSP search algorithms.

Asynchronous Partial Overlay is actually a family of algorithms. The completeness and termination problems that are presented and corrected in the present study apply to all the members of the family. The OptAPO algorithm (Mailler & Lesser, 2004; Mailler, 2004) is an optimization version of APO that solves Distributed Constraint Optimization Problems (DisCOPs). The present paper shows that similar modification to the ones made to the APO algorithm must also be applied to OptAPO in order to ensure its correctness. These changes call for performance evaluation of the resulting CompOptAPO algorithm.

The experimental evaluation that was presented in section 9 demonstrates that the performance of CompAPO is poor compared to other asynchronous search algorithms. On randomly generated DisCSPs the runtime of APO, as measured by NCCCs, is longer by up to two orders of magnitude than that of ABT (Yokoo et al., 1998; Yokoo & Hirayama, 2000) and AFC-CBJ (Meisels & Zivan, 2007).

The total number of messages sent by CompAPO is considerably smaller than the corresponding number for ABT or AFC-CBJ. This is a clear result of the fact that hard problem instances tend to be solved by a small number of mediators in a semi-centralized manner.

The runtime performance of CompOptAPO is better than that of ADOPT (Modi et al., 2005) and SyncBB (Hirayama & Yokoo, 1997) for hard instances of randomly generated DisCOPs. Similarly to the DisCSP case, the total number of messages sent by CompOptAPO is considerably smaller than the corresponding number for other DisCOP algorithms. However, in the phase-transition region of randomly generated DisCOPs, the runtime of CompOptAPO is longer by more than an order of magnitude than that of AFB (Gershman et al., 2006).

Completeness and Performance of the APO Algorithm